\documentclass[runningheads]{llncs}
\usepackage{graphicx}
\usepackage{amsmath,amssymb} 
\usepackage{color}
\usepackage{bm}
\usepackage{algorithm} 
\usepackage{algorithmic} 
\usepackage{tabularx}
\usepackage{multirow}
\usepackage{float}
\usepackage{subfigure}
\usepackage{times}
\usepackage[colorlinks,linkcolor=red]{hyperref}
\usepackage[width=122mm,left=12mm,paperwidth=146mm,height=193mm,top=12mm,paperheight=217mm]{geometry}
\begin{document}
\pagestyle{headings}
\mainmatter

\title{Superpixel-based Two-view Deterministic Fitting for Multiple-structure Data} 

\titlerunning{Superpixel-based Two-view Deterministic Fitting}

\authorrunning{G. Xiao, H. Wang, Y. Yan and D. Suter}

\author{Guobao Xiao$^{1}$, Hanzi Wang$^{1,*}$, Yan Yan$^{1}$ and David Suter$^2$}


\institute{\scriptsize{$^1$Fujian Key Laboratory of Sensing and Computing for Smart City, School of Information Science and Engineering, Xiamen University, China}\\
\scriptsize{$^2$School of Computer Science, The University of Adelaide, Australia}}

\maketitle

\begin{abstract}
This paper proposes a two-view deterministic geometric model fitting method, termed Superpixel-based Deterministic Fitting (SDF), for multiple-structure data. SDF starts from superpixel segmentation, which effectively captures prior information of feature appearances. The feature appearances are beneficial to reduce the computational complexity for deterministic fitting methods. SDF also includes two original elements, i.e., a deterministic sampling algorithm and a novel model selection algorithm. The two algorithms are tightly coupled to boost the performance of SDF in both speed and accuracy. Specifically, the proposed sampling algorithm leverages the grouping cues of superpixels to generate reliable and consistent hypotheses. The proposed model selection algorithm further makes use of desirable properties of the generated hypotheses, to improve the conventional ¡°fit-and-remove¡± framework for more efficient and effective performance. The key characteristic of SDF is that it can efficiently and deterministically estimate the parameters of model instances in multi-structure data. Experimental results demonstrate that the proposed SDF shows superiority over several state-of-the-art fitting methods for real images with single-structure and multiple-structure data.
\keywords{Deterministic algorithm, superpixel, model fitting, feature appearances.}
\end{abstract}

\section{Introduction}
\begin{figure}[t]
\centering
\centerline{\includegraphics[width=1.04\textwidth]{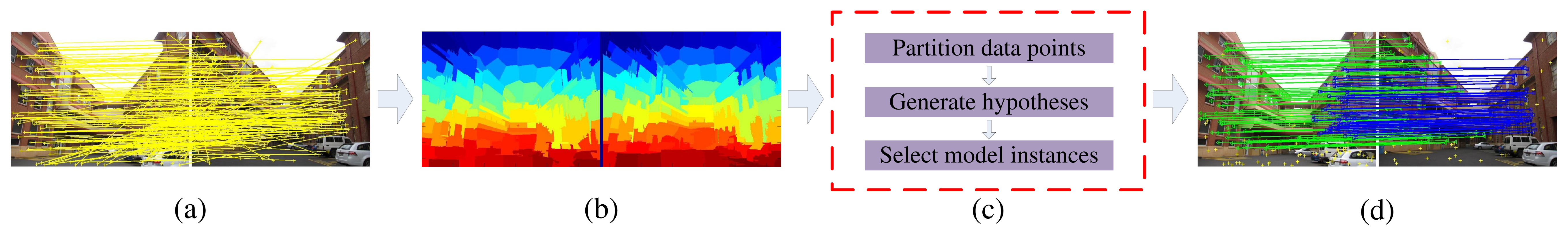}}
\vspace{-0.6cm}
\caption{Overview of the proposed method for homography estimation. (a) An image pair with keypoint correspondences. (b) Superpixel generation (each segment with the same color denotes a superpixel). (c) The procedure of the proposed method. (d) The fitting result according to the estimated model instances (the keypoint correspondences with the same color belong to the inliers of the same model instances).}
\label{fig:overview}
\end{figure}
Geometric model fitting is a challenging problem in computer vision. A major problem in model fitting is how to tolerate numerous outliers\let\thefootnote\relax\footnotetext{\scriptsize*Corresponding author}, which are ubiquitous in the real-world. RANSAC~\cite{fischler1981random} is one of the most popular fitting methods due to its robustness to outliers. Using the random sampling technique as RANSAC,  many robust fitting methods (e.g., gpbM~\cite{mittal2012generalized}, SCAMS~\cite{li2014scams}, RCG~\cite{liu2012efficient} and PEARL~\cite{pearl:ijcv12,isack2014energy}) have been proposed to improve RANSAC. There are also many robust fitting methods (e.g., SWIFT~\cite{Jaberi2015swift} and T-linkage~\cite{Magri_2014_CVPR}) developed based on different sampling techniques during the past few decades. However, these fitting methods cannot guarantee the consistency in their solutions due to their randomized nature. As a consequence, the fitting results may vary if these methods do not sample a sufficient number of subsets.

Recently, some deterministic methods (e.g., \cite{lee2013deterministic,li2009consensus,litman2015inverting,fredriksson2015practical,chin2015efficient}) have received much attention for model fitting. In contrast to the unpredictability of non-deterministic fitting methods, these deterministic fitting methods can deterministically yield solutions. For example, Li~\cite{li2009consensus} proposed to formulate the fitting problem as a mixed integer programming problem and deterministically solve the problem by using a tailored branch-and-bound scheme. Lee et al.~\cite{lee2013deterministic} proposed to employ the maximum feasible subsystem framework to deterministically generate hypotheses for the fitting problem. Litman et al.~\cite{litman2015inverting} proposed to detect a globally optimal transformation based on inlier rate estimation. Fredriksson et al.~\cite{fredriksson2015practical} proposed a branch and bound approach for the two-view translation estimation problem. Chin et al.~\cite{chin2015efficient} proposed to formulate the fitting problem as a tree search problem by which globally optimal solutions can be found based on the Astar search algorithm \cite{hart1968formal}.

Although existing deterministic fitting methods (e.g., \cite{lee2013deterministic,li2009consensus,litman2015inverting,fredriksson2015practical,chin2015efficient}) can guarantee the consistency in their solutions, most of them are computationally expensive, especially for data with few inliers. Moreover, most deterministic methods~\cite{li2009consensus,litman2015inverting,fredriksson2015practical,chin2015efficient} assume that there exists only a single model instance in data, which restricts the application of these methods in the real-world.

In this paper, we aim to solve a harder problem, where the proposed fitting method is used to efficiently and deterministically deal with multiple-structure data.  Note that feature appearances contain important prior information, and some works have been proposed to introduce feature appearances to model fitting (e.g.,~\cite{isack2014energy,chum2005matching,serradell2010combining}). However, few deterministic fitting methods fully take advantage of feature appearances of keypoint correspondences. Thus, we propose to use prior information of feature appearances to reduce the computational cost of deterministic fitting methods. More specifically, we first obtain grouping cues from superpixels, which can characterize prior information of feature appearances.
Then, based on the grouping cues, we propose an efficient and effective method, called Superpixel-based Deterministic Fitting (SDF), for multiple-structure data. The proposed SDF can deterministically generate ``high-quality" hypotheses (i.e., the hypotheses mainly include good hypotheses with only a small percentage of  bad hypotheses), and efficiently select significant hypotheses as model instances by using a novel model selection algorithm. Fig.~\ref{fig:overview} illustrates the overview of the proposed method for homography estimation.

This paper has three main contributions.  First, superpixels are introduced for deterministic model fitting. Superpixels consider spatial homogeneity, and they can provide powerful grouping cues to deterministically deal with model fitting. To the best of our knowledge, it is the first time that superpixels, which capture prior information of feature appearances, are introduced for robust model fitting in an effective manner.
Second, a deterministic sampling algorithm is proposed to exploit the grouping cues of superpixels and the corresponding keypoint matching information. With the aid of superpixels, the proposed sampling algorithm can deterministically generate high-quality hypotheses with a low percentage of bad hypotheses. Third, a novel model selection algorithm, which improves the conventional ``fit-and-remove'' framework by sequentially removing hypotheses rather than keypoint correspondences, is developed to find all model instances in data. The developed model selection algorithm is very efficient and effective since it does not require to generate new hypotheses in each iteration. Overall, SDF can efficiently and deterministically provide consistent solutions for model fitting. This is significant since most conventional fitting methods are based on randomized nature, and most existing deterministic fitting methods suffer from high computational complexity. Experimental results demonstrate that the proposed SDF can achieve substantial improvements over several recently developed state-of-the-art fitting methods.

\section{Introducing Superpixels to Model Fitting}
\label{sec:superpixel}
\begin{figure}[t]
\centering
\centerline{\includegraphics[width=0.45\textwidth]{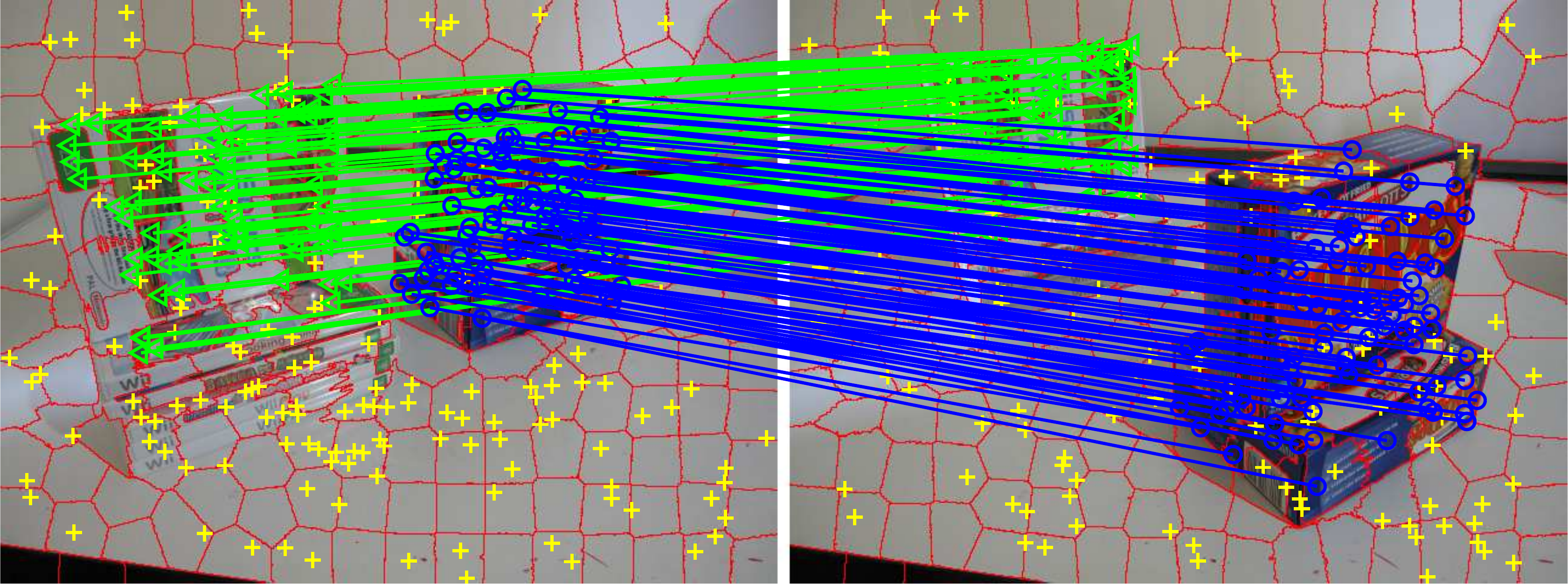}}
\vspace{-0.45cm}
\caption{An example of superpixels and keypoint correspondences based on the ground truth result of fundamental matrix for an image pair (``Gamebiscuit").}
\label{fig:supixel}
\end{figure}
In this section, we aim to obtain prior information of feature appearances, to accelerate deterministic subset sampling for hypothesis generation. The feature appearances can be derived from region consistency, which means that features within the same segments are most likely to be assigned to the same labels~\cite{kohli2009robust}. We note that superpixels obtained by an image segmentation method (such as \cite{achanta2012slic,shen2014lazy}) can adhere well to the object boundaries in the image. Moreover, one superpixel has less chance to cut across two or more objects. Thus, superpixels can be used to measure the level of consensus in labeling features for region consistency.

According to region consistency and characteristics of superpixels, two keypoint correspondences $x_i,x_j$ (a keypoint correspondence $x_i$ consists of a feature pair $\{f_i^1,f_i^2\}$ in two views) have a high possibility of belonging to the inliers of the same structure if two features $f_i^k,f_j^k$ from the $k$-th view belong to the same superpixel. 
For example, as shown in Fig.~\ref{fig:supixel} (we use the image pair ``Gamebiscuit" from the AdelaideRMF datasets \cite{wong2011dynamic}), we perform superpixel segmentation~\cite{achanta2012slic} on the image pair and show keypoint correspondences based on the ground truth result of fundamental matrix. We can see that most keypoint correspondences, whose valid features in one view come from the same superpixels, belong to the same structures.

Inspired by the above observations, the keypoint correspondences can be partitioned into a set of groups (i.e., ${\bm{\mathcal{G}}}$) (where each group consists of the keypoint correspondences associated to the features within the same superpixel), with the intuition that the keypoint correspondences of a group have a high possibility of belonging to the same structure. This will help to accelerate deterministic subset sampling for hypothesis generation.

\begin{figure}[t]
\centering
\centerline{\includegraphics[width=0.45\textwidth]{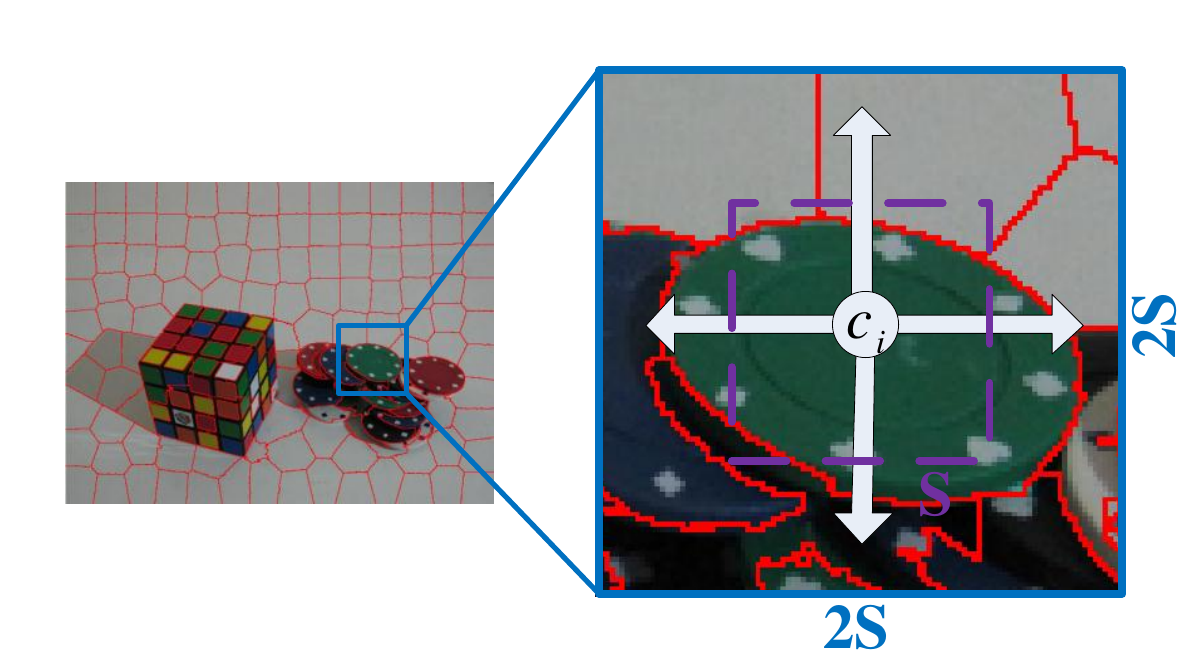}}
\vspace{-0.5cm}
\caption{An example of group combination. $c_i$ is the center of the $i$-{th} group in $\bm{\mathcal{G}}$ and $S$ is the grid interval. The purple dashed box denotes a superpixel size and we perform the combining procedure within a $2S\times2S$ region (i.e., the blue solid box).}
\label{fig:limitedregion}
\end{figure}
\section{The Proposed Method}
\label{sec:fittingmethod}
In this section, based on the prior information of feature appearances derived from superpixels (discussed in Sec.~\ref{sec:superpixel}), we first present a deterministic sampling algorithm for hypothesis generation in Sec.~\ref{sec:hypothesesgeneration}. Then, in Sec.~\ref{sec:modelselection}, we present a novel ``fit-and-remove'' framework for model selection. Finally, we summarize the complete fitting method in Sec.~\ref{sec:completedmethod}.
\vspace{-0.5cm}
\subsection{Hypothesis Generation}
\label{sec:hypothesesgeneration}
\vspace{-0.2cm}
The prior information of feature appearances can provide powerful grouping cues for hypothesis generation. However, grouping cues cannot be directly used to generate hypotheses. This is caused by two problems: (i) The inliers of a structure in data often include more than the keypoint correspondences in a group, which may cause degeneration in sampled subsets with small spans~\cite{tran2014sampling}; (ii) The keypoint correspondences of a group may contain outliers, which will lead to the failure of estimating a true structure in data.

We propose two strategies to alleviate the above-mentioned problems: For the first problem, we increase the spans of sampled subsets by combining groups in an input group set $\bm{\mathcal{G}}$. Theoretically, we can combine any two groups in $\bm{\mathcal{G}}$, but such a strategy is time consuming and will also generate a large number of ``bad" groups that consist of keypoint correspondences belonging to different structures. Thus, we propose to only combine groups within a limited region, since the keypoint correspondences of neighboring groups, have a high possibility of belonging to the same structure. More specifically, for a group $\mathcal{G}_i$$\in$$ \bm{\mathcal{G}}$, we combine it with each of its neighboring group within a limited region to generate a new group $\hat{\mathcal{G}}_{i\cup j}$:
\begin{align}
 \label{equ:combine}
\hat{\mathcal{G}}_{i\cup j}&=\left\{ \begin{array}    {r@{\quad \quad} l}
\mathcal{G}_i\cup \mathcal{G}_j, & if~\mathcal{G}_j\in \bm{\mathcal{N}(}\mathcal{G}_i\bm)~and~R(l_i,l_j)\leq 2S\times2S,\\
\mathcal{G}_i~~~~,&otherwise,
\end{array}\right.
\end{align}
where $\bm{\mathcal{N}(}\mathcal{G}_i\bm)$ is the neighboring group of $\mathcal{G}_i$. $l_i$ and $l_j$ denote the corresponding superpixels of  $\mathcal{G}_i$ and $\mathcal{G}_j$, respectively. $R(.,.)$ denotes the combined region of two superpixels in the image. Here, according to the expected superpixel size ($S$$\times $$S$), we compute the grid interval $S$ as~\cite{achanta2012slic}, i.e., $S$$=$$\sqrt{N/M}$, where $N$ and $M$ are the number of pixels and superpixels, respectively. An example of group combination is illustrated in Fig.~\ref{fig:limitedregion}.

In this manner, we can obtain a set of combined groups $\bm{\hat\mathcal{G}}$, where small-size groups are combined, while large-size groups (whose sizes are larger than $2S$$\times$$2S$) are not combined. Despite some combined groups may consist of keypoint correspondences belonging to two different structures, the degeneration problem can be effectively alleviated since each group $\hat{\mathcal{G}}_{i}$ in $\bm{\hat\mathcal{G}}$ includes keypoint correspondences with larger spans. 

For the second problem, we only consider the most ``promising" keypoint correspondences in a combined group $\hat{\mathcal{G}}_{i}$$=$$\{x_{i}^j\}_{j=1}^{n_i}$ according to the corresponding matching information. Specifically, for a group $\hat{\mathcal{G}}_{i}$, {by sorting keypoint correspondences according to the corresponding matching score vector $\bm{s_i}$$=$$[s_i^1~s_i^2~\cdots~s_i^{n_i}]$ (each score $s_i^j$ is computed according to the SIFT correspondences~\cite{lowe2004distinctive}), we can find a permutation}:
\begin{align}
\label{equ:sorting2}
\bm{a_i}=[a_i^1~a_i^2~\cdots~a_i^{n_i}],
\end{align}
where $a_i^j$ is the ranking index of the $j$-th keypoint correspondence in the $i$-th group $\hat{\mathcal{G}}_{i}$. The keypoint correspondences in $\hat{\mathcal{G}}_{i}$ are sorted in the non-ascending order, i.e.,
\begin{align}
 \label{equ:sorting}
u<v \Longrightarrow s_i^{a_i^u} > s_i^{a_i^v},
\end{align}
where $u$ and $v$ respectively denote the indices of $x_{i}^u$ and $x_{i}^v$ in $\hat{\mathcal{G}}_{i}$.

Then, for a group $\hat{\mathcal{G}}_{i}$, we sample the $m_0$ top sorted keypoint correspondences, i.e., $\{x_{i}^j\}_{j=a_i^1}^{a_i^{m_0}}$. Here, we only sample the keypoint correspondences with high matching scores to reduce the influence of outliers. This is because a keypoint correspondence with a high matching score has a higher probability to be an inlier of a structure in data \cite{brahmachari2013hop}.


Based on the prior information of feature appearances and the two above-mentioned strategies (i.e., the group combination and the promising keypoint correspondences selection), we propose a deterministic sampling algorithm for hypothesis generation (see Algorithm \ref{alg:sampling}). The proposed sampling algorithm considers both feature appearances and geometric information, and it is also tractable due to its deterministic nature. Therefore, the proposed sampling algorithm can generate reliable and consistent hypotheses for model fitting. For the parameter $m_0$, i.e., the number of keypoint correspondences we use to generate a hypothesis, we can set it as $p+2$, where $p$ denotes the minimum size of sampled subsets for computing a unique hypothesis. This is because that the sampled subset with $p+2$ keypoint correspondences can generate a stable hypothesis, which has been demonstrated in \cite{tennakoon2015robust}. It is also worth pointing out that the way we sample subsets in the local region will not affect the quality of model hypotheses, because the grouping cues include information of all keypoint correspondences, and the group combination process also allows to sample in a larger region.

\begin{algorithm}[t] 
\small
\renewcommand{\algorithmicrequire}{\textbf{Input:}}
\renewcommand\algorithmicensure {\textbf{Output:} }
\caption{The proposed deterministic sampling algorithm for hypothesis generation} 
\label{alg:sampling} 
\begin{algorithmic}[1] 
\REQUIRE 
a set of groups of keypoint correspondences ${\bm{\mathcal{G}}}$
\STATE Combine each group $\mathcal{G}_i$$\in$$\bm{\mathcal{G}}$ with each one $\mathcal{G}_j$ of its neighbors $\bm{\mathcal{N}(}\mathcal{G}_i\bm)$ within a limited region to generate a new group $\hat{\mathcal{G}}_{i\cup j}$ by Eq.~(\ref{equ:combine}).
\STATE Sort keypoint correspondences in each combined group by Eq.~(\ref{equ:sorting}).
\STATE Select the $m_0$ top sorted keypoint correspondences in the combined group as a sampled subset, which is used to generate a hypothesis $\theta_i$.
\ENSURE The generated hypothesis set {$\bm{\theta}$} (=$\{\theta_i\}_{i=1,2,\ldots}$)
\end{algorithmic}
\end{algorithm}
We note that PROSAC~\cite{chum2005matching} also employs the most promising keypoint correspondences (measured by the matching information) to generate hypotheses. However, for each hypothesis, the proposed sampling algorithm only selects the most promising keypoint correspondences from a group based on superpixels, which is more ``local" than PROSAC (recall that PROSAC samples a subset from all keypoint correspondences). That will help the proposed sampling algorithm to efficiently sample all-inlier subsets, which is more obvious on multiple-structure data. Moreover, PROSAC cannot guarantee the consistency of hypotheses due to its randomized nature. In contrast, the proposed sampling algorithm is a deterministic sampling algorithm.
\vspace{-0.5cm}
\subsection{Model Selection}
\label{sec:modelselection}
\vspace{-0.2cm}
Given the hypothesis set $\bm{\theta}$ generated by Algorithm~\ref{alg:sampling}, the next step is to select model instances for model fitting. For single-structure data, the hypothesis with the largest number of inliers is directly selected as the estimated model instance.

For multiple-structure data, we propose a novel ``fit-and-remove" framework, which sequentially selects a hypothesis $\theta$ with the largest number of inliers from the hypothesis set, and updates the hypothesis set $\bm{\theta}$ by removing some redundant hypotheses.

In contrast to the traditional ``fit-and-remove" framework, the proposed framework removes hypotheses rather than keypoint correspondences and it does not require the generation of new hypotheses during each step. Therefore, it effectively overcomes the limitations of the conventional ``fit-and-remove" framework~\cite{wang2012simultaneously}, i.e., inaccurate inlier/outlier dichotomy can lead to wrong estimation of the remaining model instances, and repeated hypothesis generation during each step is computationally inefficient as well.

\begin{figure}[t]
\centering
\begin{minipage}[t]{.22\textwidth}
  \centerline{\includegraphics[width=1.16\textwidth]{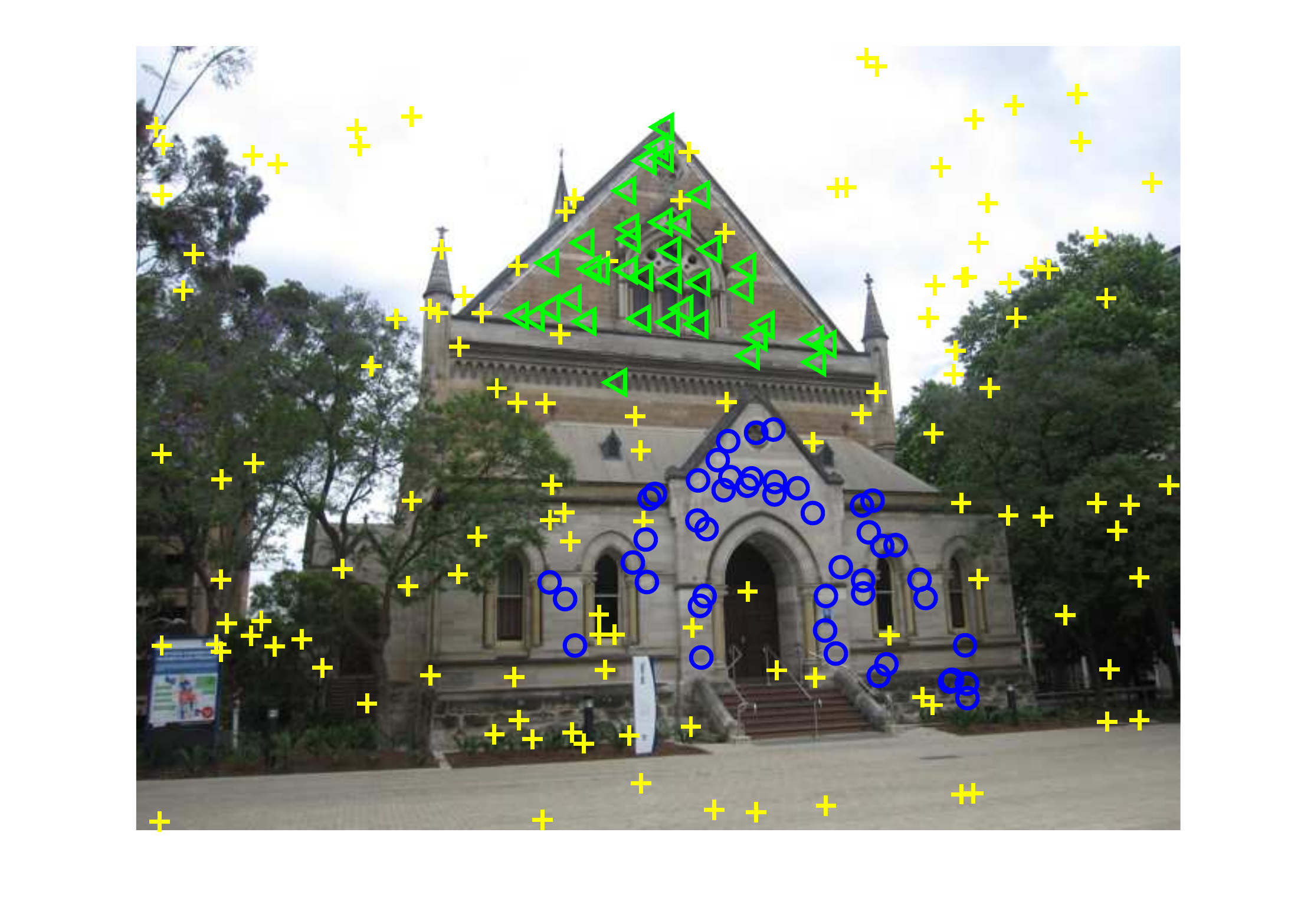}}
  \vspace{-0.35cm}
 \centerline{(a)}
  \centerline{\includegraphics[width=1.16\textwidth]{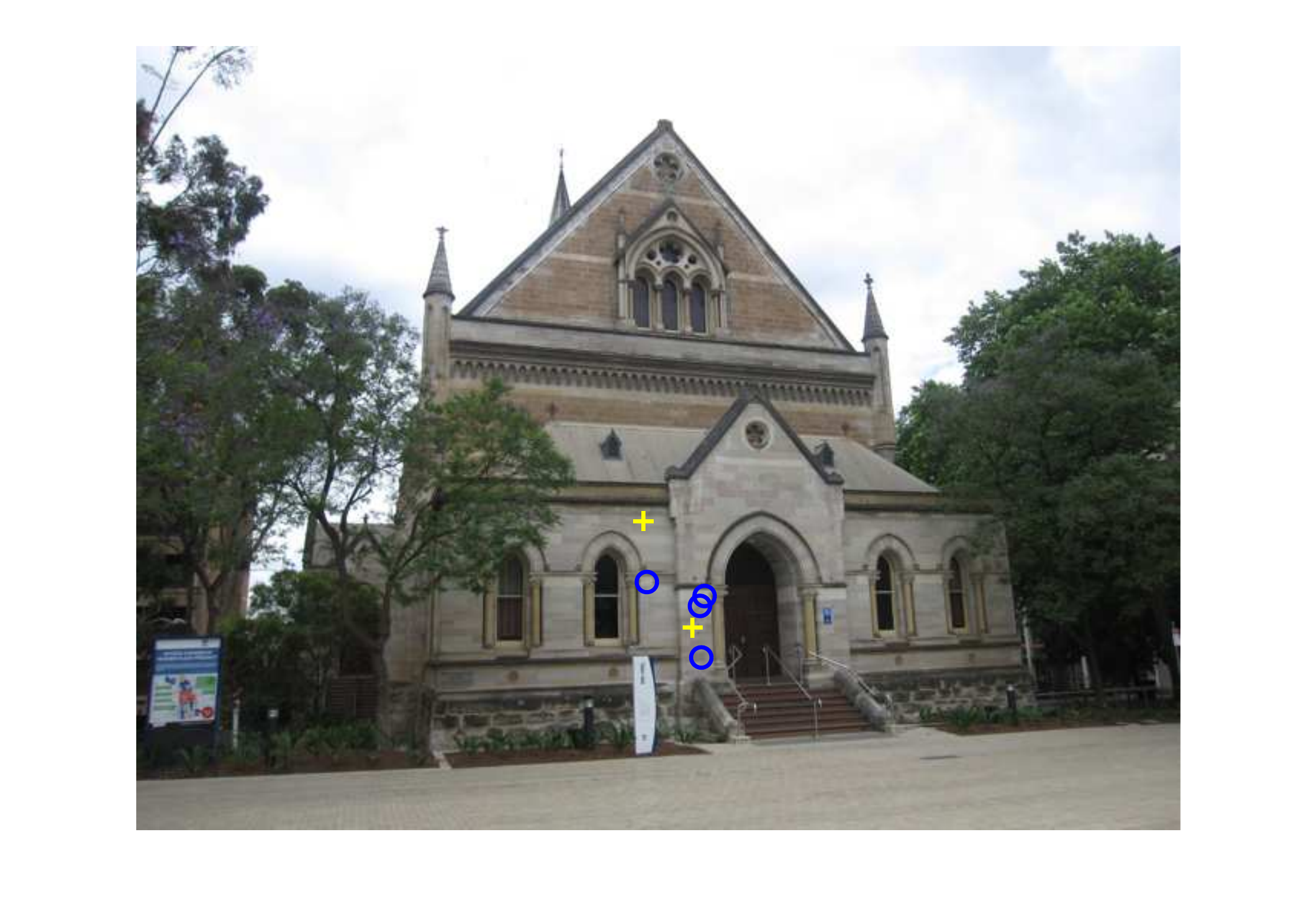}}
    \vspace{-0.35cm}
 \centerline {(d)}
\end{minipage}
\begin{minipage}{.05\textwidth}
$\qquad$
\end{minipage}
\begin{minipage}[t]{.22\textwidth}
  \centerline{\includegraphics[width=1.16\textwidth]{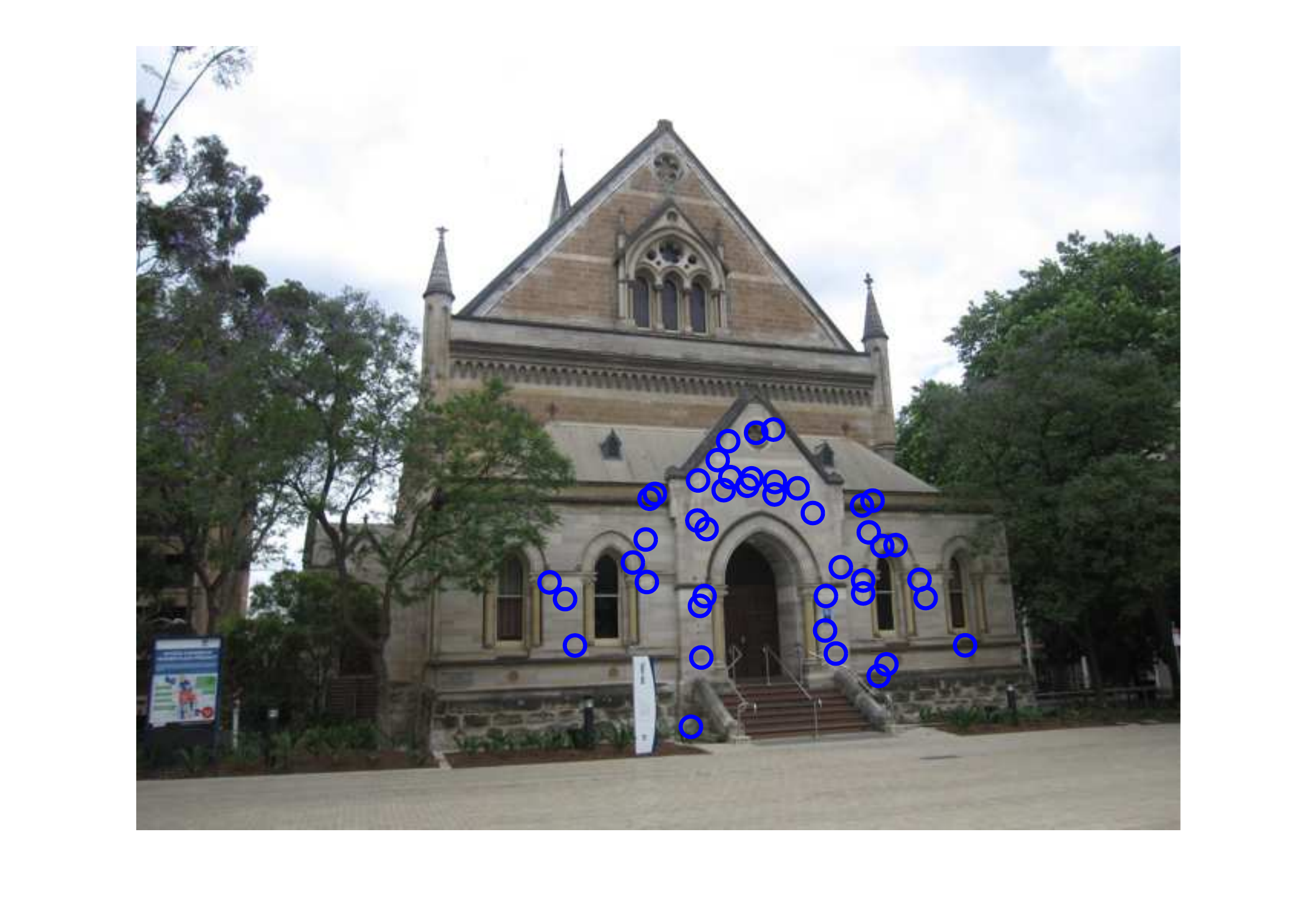}}
   \vspace{-0.35cm}
 \centerline{(b)}
    \centerline{\includegraphics[width=1.16\textwidth]{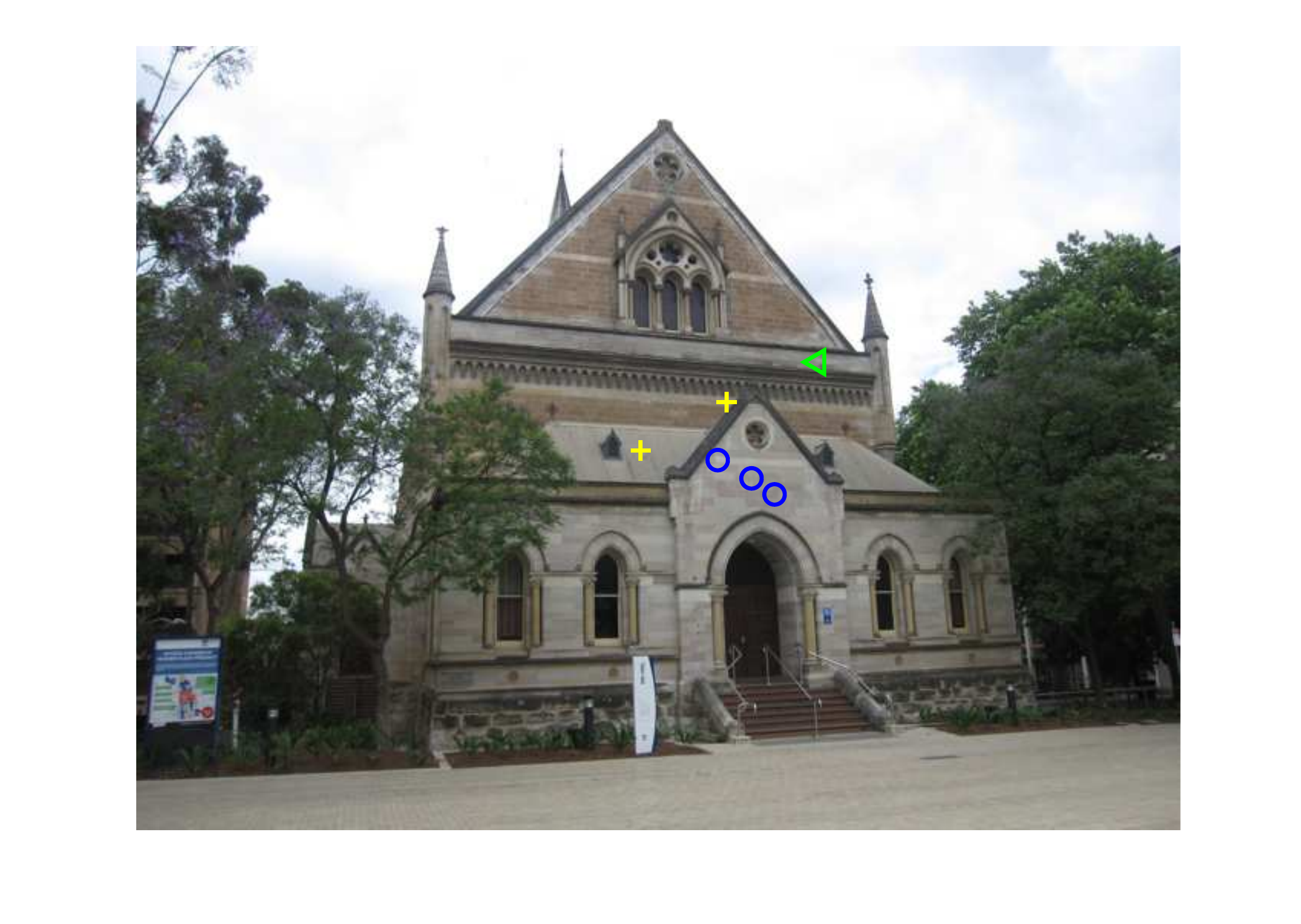}}
      \vspace{-0.35cm}
   \centerline {(e)}
\end{minipage}
\begin{minipage}{.05\textwidth}
$\qquad$
\end{minipage}
\begin{minipage}[t]{.22\textwidth}
 \centerline{\includegraphics[width=1.16\textwidth]{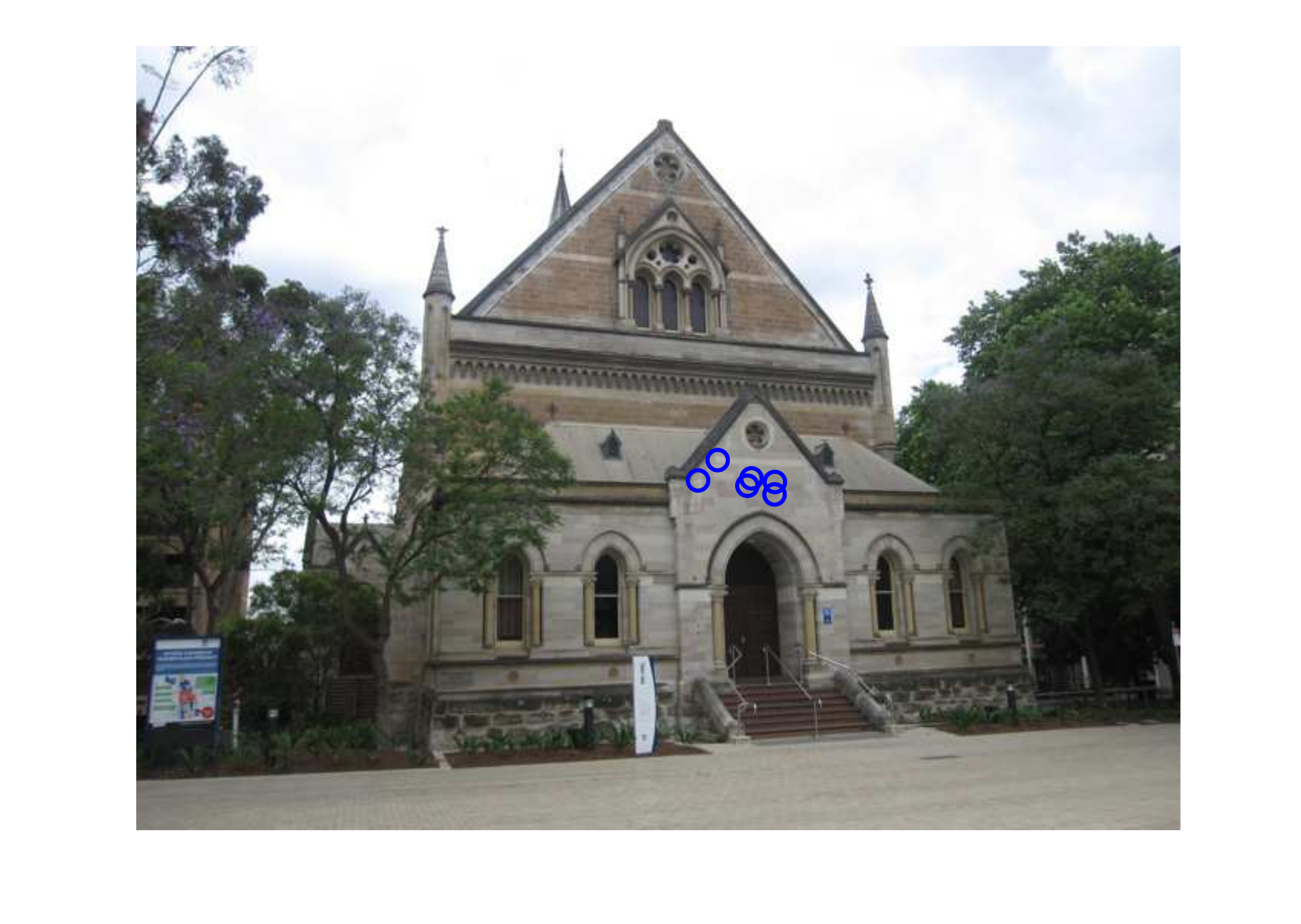}}
   \vspace{-0.35cm}
  \centerline {(c)}
  \centerline {\includegraphics[width=1.16\textwidth]{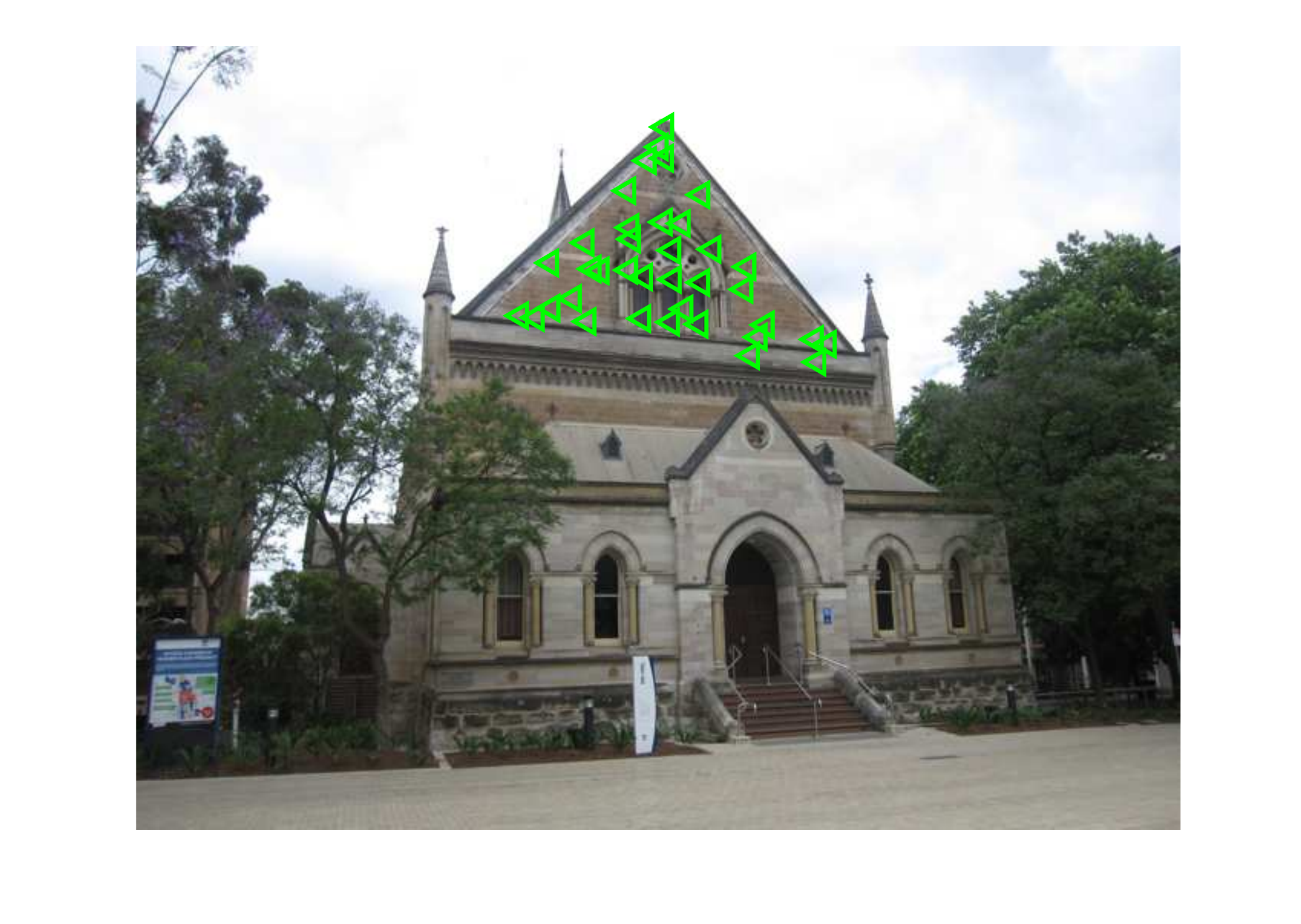}}
   \vspace{-0.35cm}
  \centerline{(f)}
\end{minipage}
  \vspace{-0.35cm}
\caption{An example of model selection for homography estimation (only one of the two views is shown): (a) An input image (``Elderhalla'') with the ground truth results. (b) The inliers of the first selected hypothesis. (c)$\sim$(e) The sampled subsets of three redundant hypotheses. (f) The inliers of the second selected hypothesis.}
\label{fig:modelselection}
\end{figure}

The key step of the proposed framework is how to remove redundant hypotheses. For a selected hypothesis $\theta_i$ (e.g., Fig.~\ref{fig:modelselection}(b)), redundant hypotheses contain bad hypotheses (e.g., Figs.~\ref{fig:modelselection}(d) and~\ref{fig:modelselection}(e)), whose sampled subset consists of outliers or keypoint correspondences from different model instances, and good hypotheses (e.g., Fig.~\ref{fig:modelselection}(c)), which correspond to the same model instance as $\theta_i$. Therefore, for the selected hypothesis $\theta_i$, we define $h(i,j)$ to determine if a hypothesis $\theta_j$ is redundant:
\begin{align}
 \label{equ:reduant}
h(i,j)&=\left\{ \begin{array}    {r@{\quad \quad} l}
1, & if~\mathbf{Sam}(\theta_j)\cap \mathbf{In}(\theta_i) \neq \emptyset,\\
0,&otherwise,
\end{array}\right.
\end{align}
where $\mathbf{Sam}(\theta_j)$ is the sampled subset of $\theta_j$ and $\mathbf{In}(\theta_i)$ is the inlier set of $\theta_i$. $\mathbf{Sam}(\theta_j)\cap \mathbf{In}(\theta_i)$ is used to decide if the sampled subset corresponding to $\theta_j$ contains any keypoint correspondence belonging to the inliers of $\theta_i$. Thus, each hypothesis $\theta_j$ with $h(i,j)$$=$$1$ will be treated as a redundant hypothesis and removed from the hypothesis set $\bm{\theta}$.

Note that the proposed framework may fail to remove redundant hypotheses based on some conventional sampling algorithms, e.g., \cite{fischler1981random,Magri_2014_CVPR}. This is because these sampling algorithms will generate a large percentage of bad hypotheses and the sampled subsets consist randomly selected keypoint correspondences. In contrast, the proposed framework can work well based on the high-quality hypotheses provided by the proposed sampling algorithm (Algorithm~\ref{alg:sampling}). Specifically, for a selected hypothesis $\theta_i$, the remaining good hypotheses corresponding to model instances in data include two parts, i.e., the hypotheses $\bm{\widehat{\theta}_i}$ corresponding to the same model instance as $\theta_i$, and the hypotheses $\bm{\widetilde{\theta}_i}$ corresponding to the remaining model instances in data. For each iteration, the proposed framework can remove $\bm{\widehat{\theta}_i}$ while preserving $\bm{\widetilde{\theta}_i}$. That is, for a hypothesis belonging to $\bm{\widehat{\theta}_i}$, it can be effectively removed because the keypoint correspondences from its sampled subset have high matching scores and most of these keypoint correspondences more likely belong to the inliers of $\theta_i$. Thus, according to Eq.~(\ref{equ:reduant}), the hypothesis is one of the redundant hypotheses. In contrast, for a hypothesis belonging to $\bm{\widetilde{\theta}_i}$, the keypoint correspondences from its sampled subset have a low probability to be the inliers of $\theta_i$ and it will not be removed.

\vspace{-0.5cm}
\subsection{The Complete Method}
\label{sec:completedmethod}
\vspace{-0.2cm}
We summarize the proposed SDF method with all the ingredients developed in the previous sections (see Algorithm~\ref{alg:completed}). We first generate superpixels of an image pair with a selected segmentation algorithm. Here, we perform the SLIC segmentation algorithm \cite{achanta2012slic} (which deterministically generates superpixels by using a variant of the $k$-means clustering algorithm) to obtain superpixels due to its simplicity and effectiveness. Moreover, it can adhere well to the object boundaries in an image with $O(N)$ complexity (where $N$ is the number of pixels). It is worth pointing out that the performance of the proposed method does not greatly depend on the quality of superpixel segmentation. This is because a model instance in data often corresponds to two or more hypotheses based on the grouping cues derived from different superpixels, and the model instance can be estimated from these hypotheses.

The proposed SDF exploits the grouping cues of superpixels to deterministically estimate the parameters of model instances in multi-structure data. SDF includes two main parts, i.e., a deterministic sampling algorithm and a novel ``fit-and-remove" framework for model selection. The proposed deterministic sampling algorithm effectively introduces feature appearances (derived from superpixels) to geometric model fitting for hypothesis generation, and the proposed ``fit-and-remove" framework takes advantages of the generated high-quality hypotheses. Therefore, the proposed sampling algorithm and the proposed model selection framework are nicely coupled, and they jointly lead to deterministic fitting results. The computational complexity of SDF is approximately proportional to $O(N)$. Among all the steps of the proposed SDF, the step of superpixel segmentation (i.e., step~\ref{state:perfomsuperpixel}) consumes the majority of the computational time of SDF.

Note that GroupSAC~\cite{ni2009groupsac} also partitions keypoint correspondences into a set of groups. However, the groups partitioned by the proposed SDF are smaller and more accurate than those partitioned by GroupSAC due to the over-segmentation nature in superpixels. In addition, GroupSAC only works for single-structure data with randomized nature, and its performance greatly depends on the quality of image segmentation. In contrast, SDF has significant superiority over GroupSAC since SDF can deterministically deal with multi-structure data.
\begin{algorithm}[t] 
\small
\renewcommand{\algorithmicrequire}{\textbf{Input:}}
\renewcommand\algorithmicensure {\textbf{Output:} }
\caption{The superpixel-based two-view deterministic fitting method} 
\label{alg:completed} 
\begin{algorithmic}[1] 
\REQUIRE 
keypoint correspondences, the inlier scale and the number of model instances $T$
\STATE Perform the superpixel segmentation algorithm \cite{achanta2012slic} on a tested image pair.
\label{state:perfomsuperpixel}
\STATE Partition keypoint correspondences into a set of groups $\bm{\mathcal{G}}$ based on the segmented superpixels (described in Sec.~\ref{sec:superpixel}).
\label{state:partition}
\STATE Deterministically generate hypotheses $\bm{\theta}$ by Algorithm~\ref{alg:sampling}.
\label{state:gereratehypotheses}
\FOR{$i=1$ to $T$}
\label{state:modelselection}
\STATE Select a hypothesis $\theta_i$ with the largest number of inliers (based on the input inlier scale) from $\bm{\theta}$ as an estimated model instance.
\STATE Find the redundant hypotheses $\bm{\vartheta}_i$($=\{\theta_i^j\}_{j=1,2,\ldots}$) with respect to the selected hypothesis $\theta_i$ according to Eq.~(\ref{equ:reduant}).
\STATE Remove the selected hypothesis $\theta_i$ and the redundant hypotheses $\bm{\vartheta}_i$ from $\bm{\theta}$, i.e., $\bm{\theta}\leftarrow \bm{\theta}\setminus \{\bm{\vartheta}_i\cup \theta_i\}$.
\ENDFOR
\label{state:modelselection2}
\ENSURE The parameters of estimated model instances
\end{algorithmic}
\end{algorithm}
\section{Experiments}
\label{sec:experiments}
In this section, we perform homography estimation and fundamental matrix estimation on single-structure and multiple-structure datasets. We compare the proposed SDF with several state-of-the-art model fitting methods, including PROSAC~\cite{chum2005matching}, AStar~\cite{chin2015efficient} and T-linkage~\cite{Magri_2014_CVPR}. PROSAC is evaluated since it also considers feature appearances as SDF. AStar is one of the state-of-the-art methods for deterministic fitting. However,  AStar only works on single-structure data and thus we do not evaluate it in subsection~\ref{sec:singlestructure}. T-linkage is a representative model fitting method that effectively works on multiple-structure data, but it can not work on single-structure data very well due to the used outlier rejection process. Thus we only use it as the competing method on multiple-structure datasets in Section~\ref{sec:multiplestructures}. In addition, we also run RANSAC as a baseline. 

For the parameter settings, we use the same inlier scale on each dataset for all the competing methods and also optimize the other parameters of all the competing methods on each dataset for the best performance. All experiments are run on MS Windows $7$ with Intel Core i$7$-$3630$ CPU $2.4GHz$ and $16GB$ RAM.

\begin{figure}[t]
\centering
\begin{minipage}[t]{.4\textwidth}
  \centering
  \centerline{\includegraphics[width=0.98\textwidth]{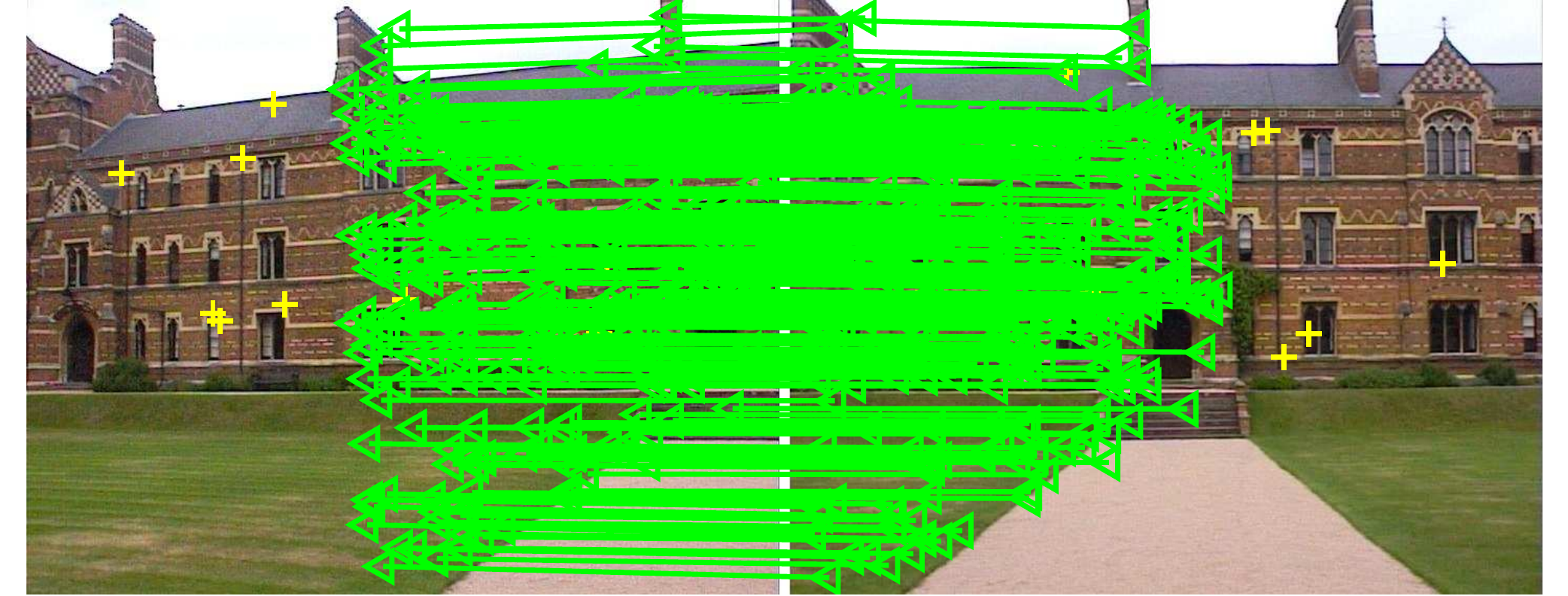}}
     \vspace{-0.1cm}
  \centerline{(a) Keble}
  \centerline{\includegraphics[width=0.98\textwidth]{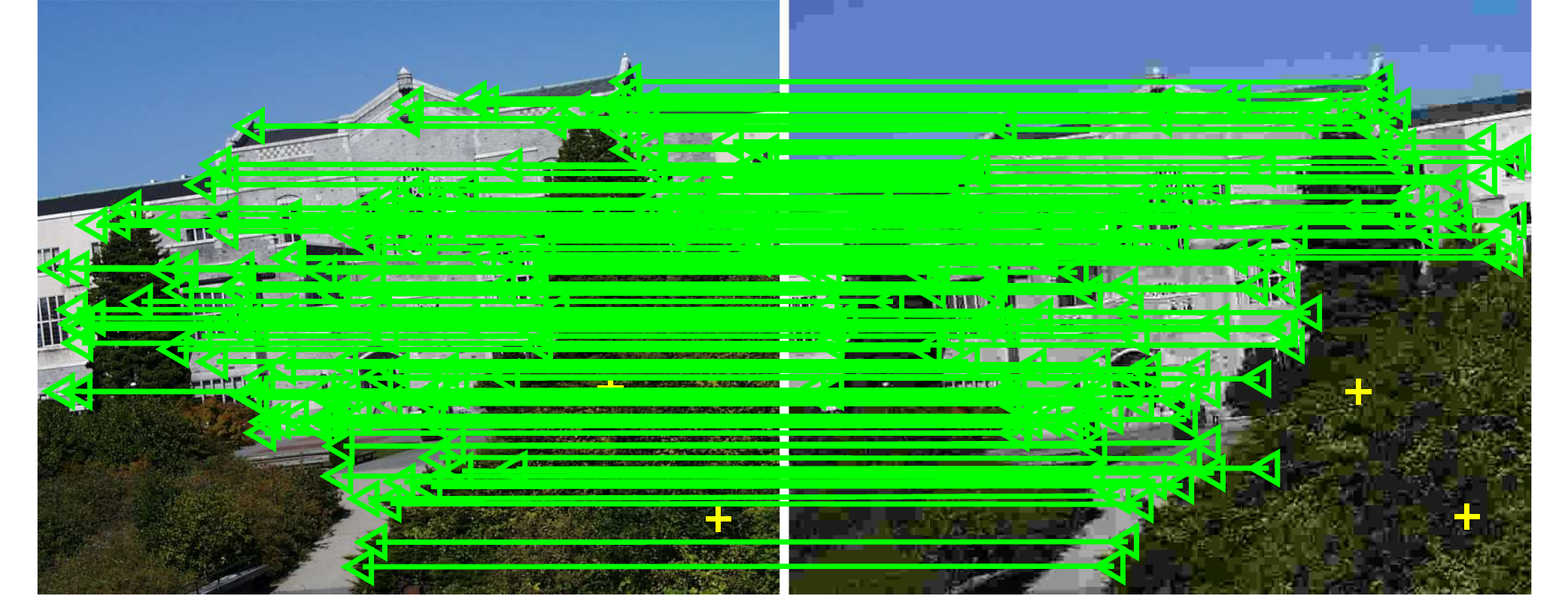}}
     \vspace{-0.1cm}
    \centerline{(b) Ubc}
    \centerline{\includegraphics[width=0.98\textwidth]{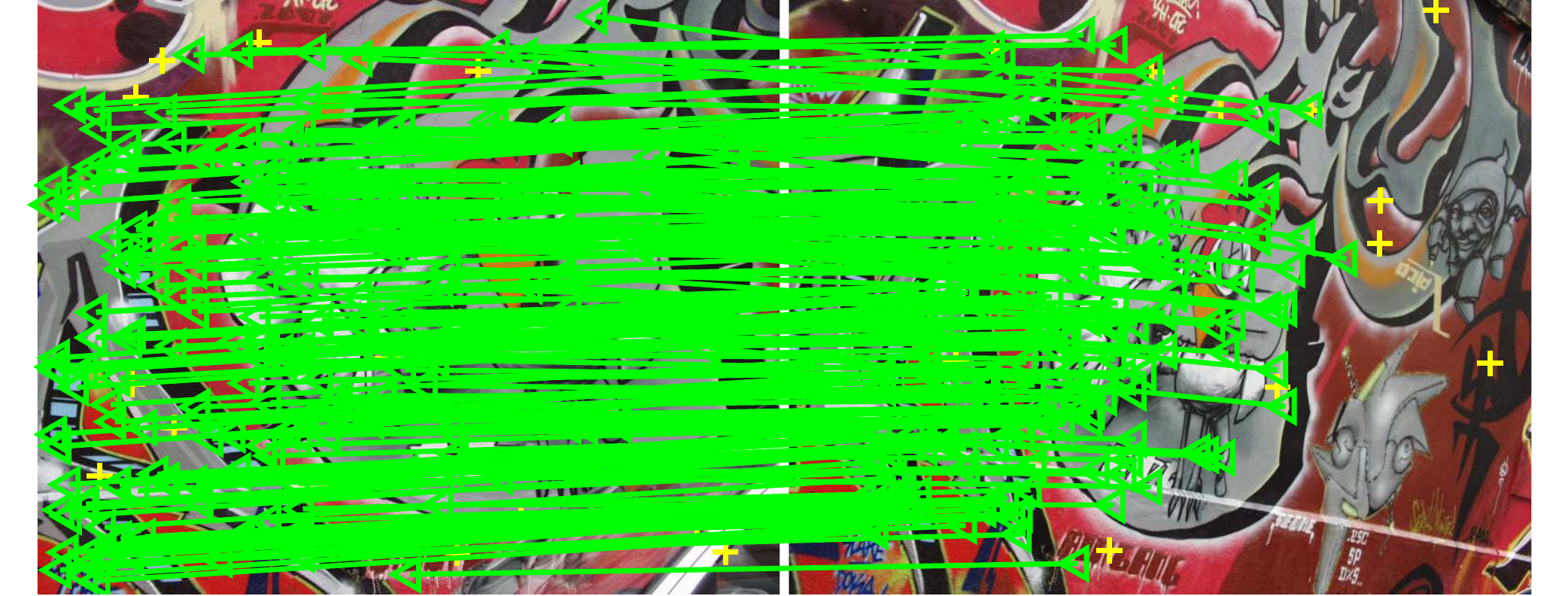}}
        \vspace{-0.1cm}
  \centerline{(c) Graffiti}
  \centerline{\includegraphics[width=0.98\textwidth]{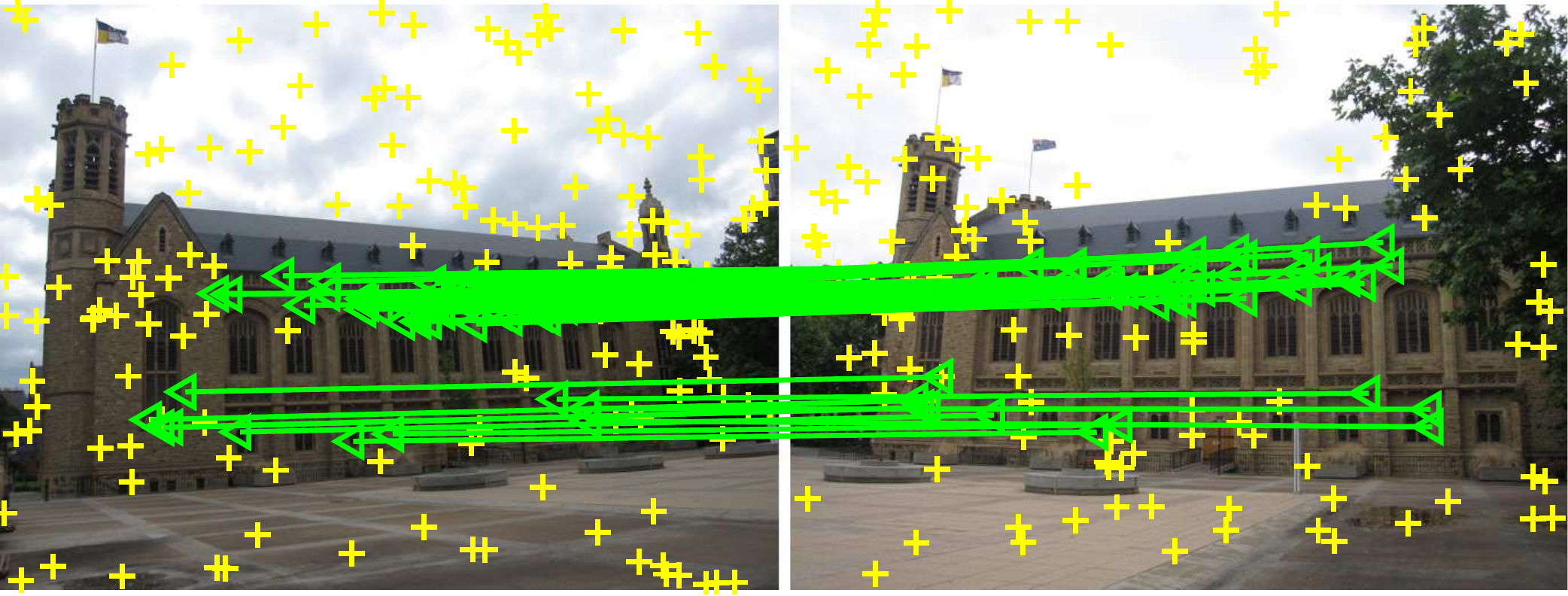}}
       \vspace{-0.1cm}
    \centerline{(d) Bonython}
      \centerline{\includegraphics[width=0.98\textwidth]{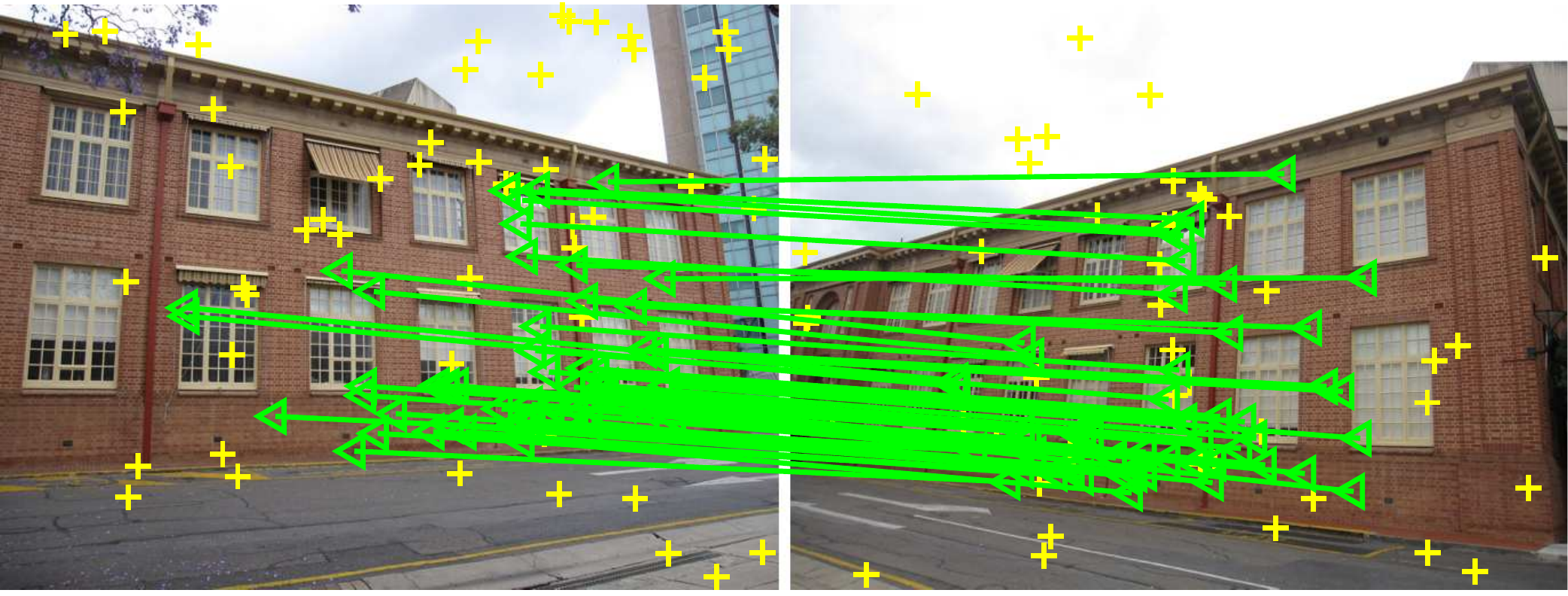}}
       \vspace{-0.1cm}
        \centerline{(e) Physics}
\end{minipage}
\begin{minipage}{.1\textwidth}
$\qquad$
\end{minipage}
\begin{minipage}[t]{.4\textwidth}
  \centering
  \centerline{\includegraphics[width=0.98\textwidth]{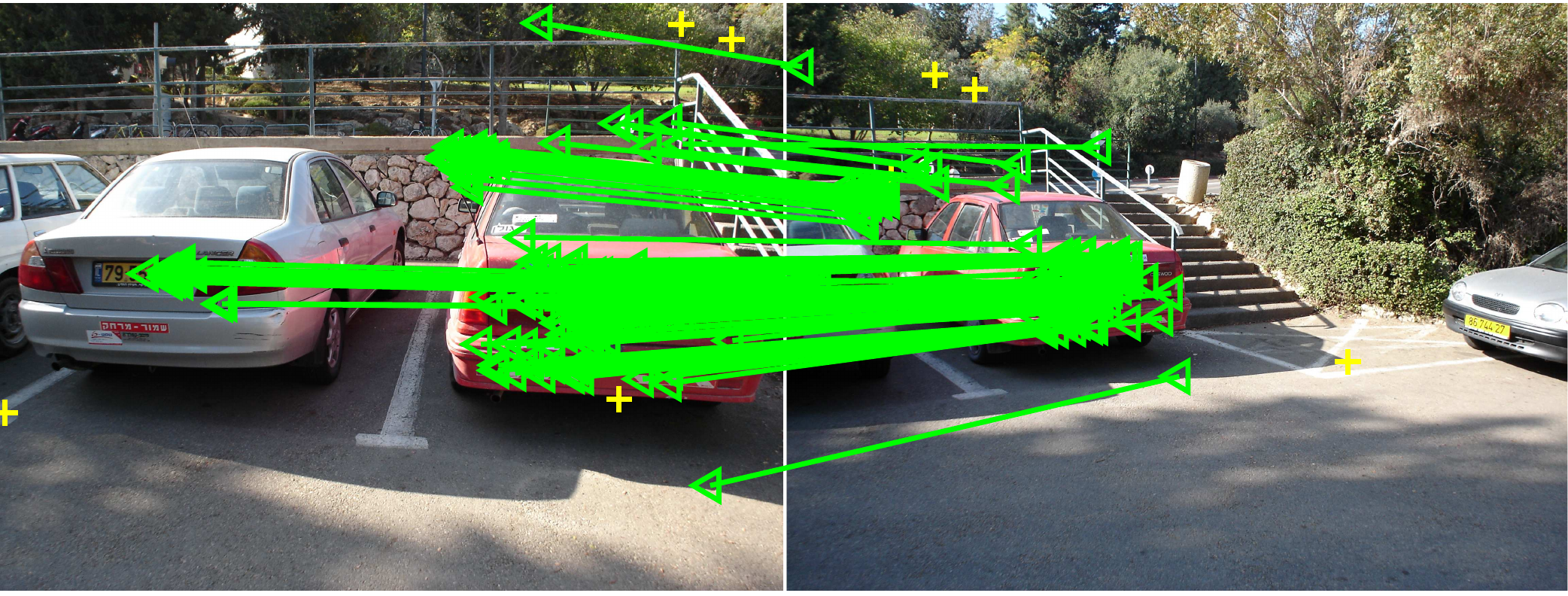}}
    \vspace{-0.1cm}
  \centerline{(f) Twocars }
  \centerline{\includegraphics[width=0.98\textwidth]{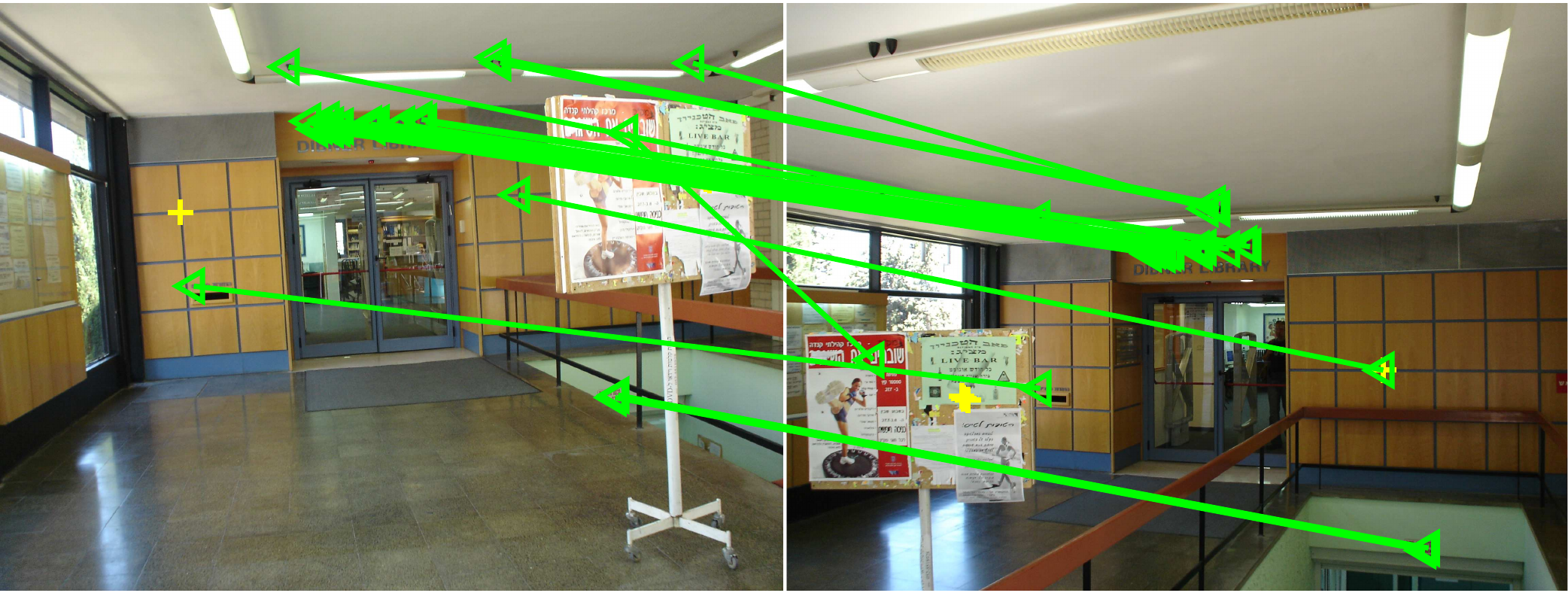}}
     \vspace{-0.1cm}
  \centerline{(g) Library}
  \centerline{\includegraphics[width=0.98\textwidth]{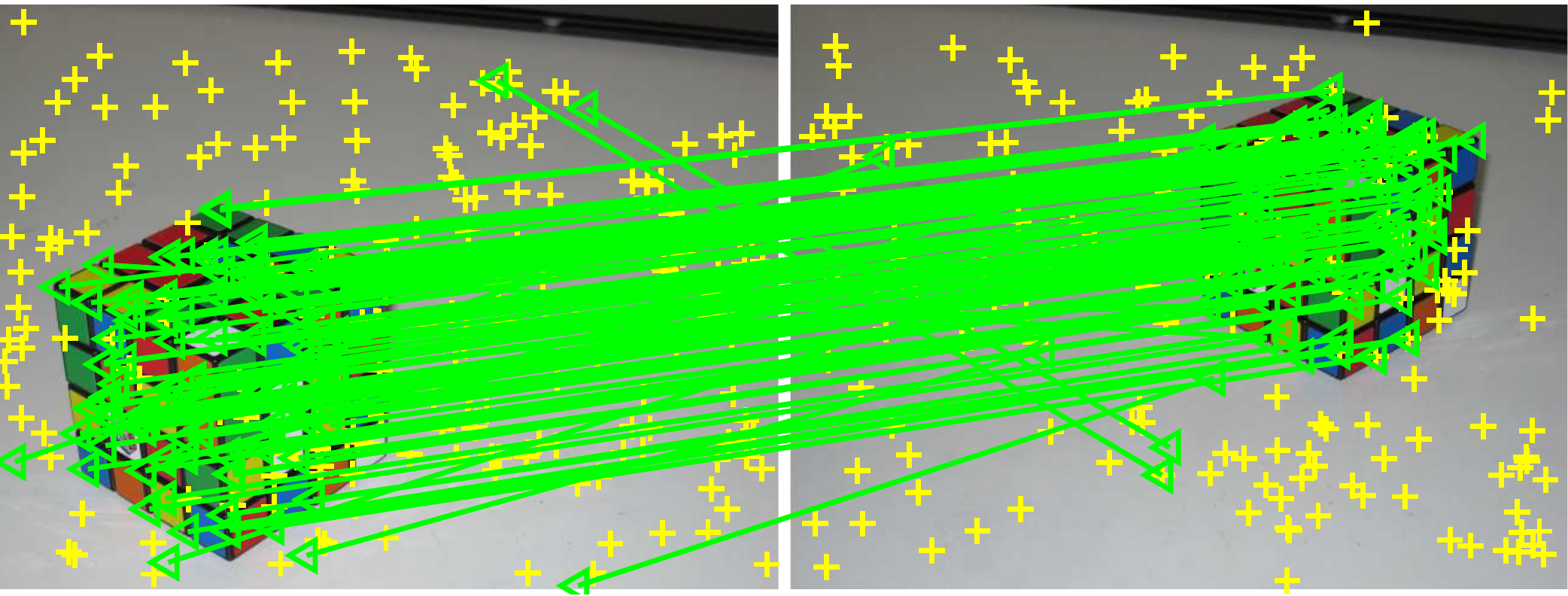}}
     \vspace{-0.1cm}
  \centerline{(h) Cube}
  \centerline{\includegraphics[width=0.98\textwidth]{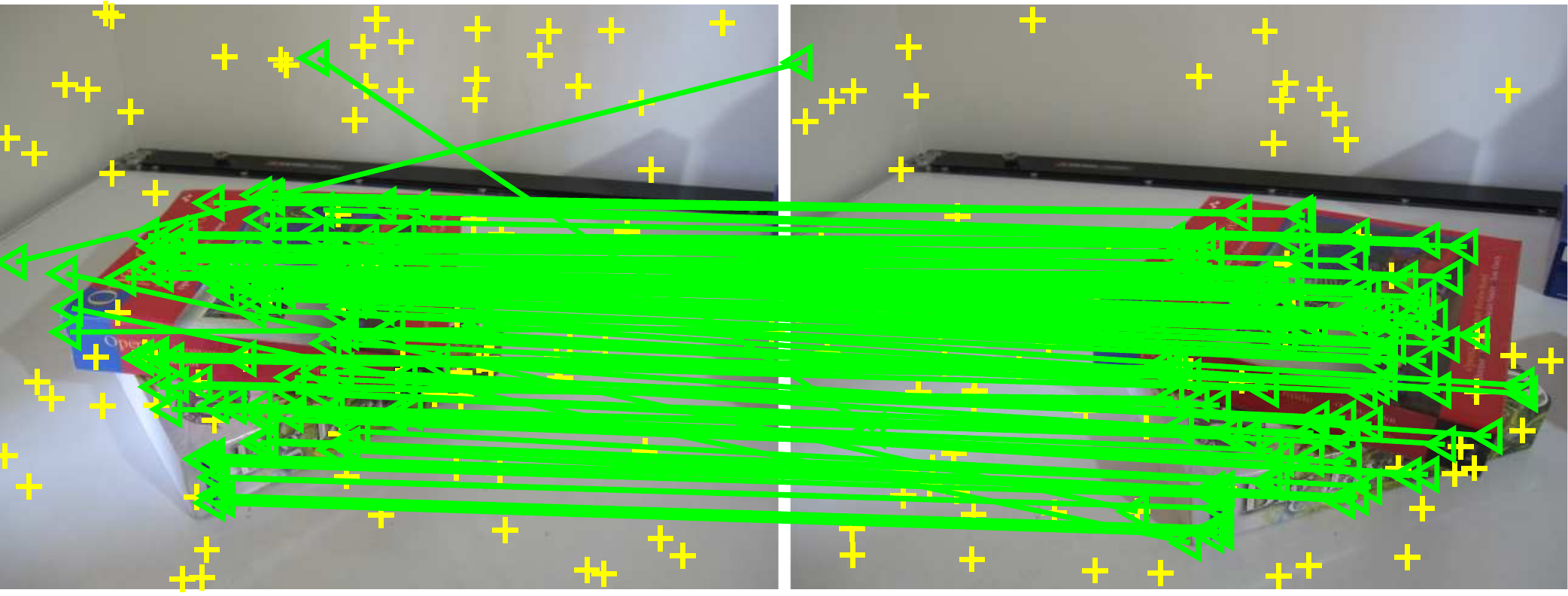}}
       \vspace{-0.1cm}
    \centerline{(i) Book}
      \centerline{\includegraphics[width=0.98\textwidth]{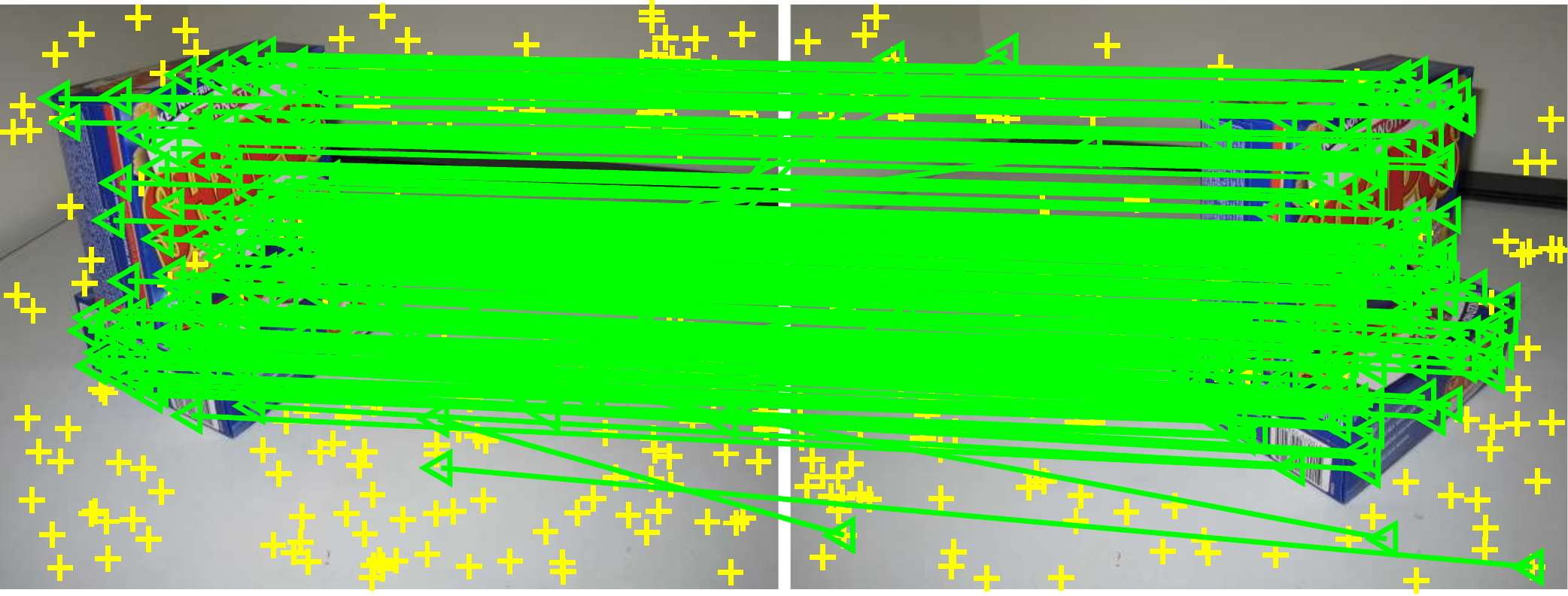}}
       \vspace{-0.1cm}
      \centerline{(j) Biscuit}
\end{minipage}
   \vspace{-0.35cm}
\caption{Fitting results obtained by SDF on 10 image pairs with single-structure data. (a) $\sim$ (e) show the results obtained by SDF on homography estimation, and (f) $\sim$ (j) show the results obtained by SDF on fundamental matrix estimation. We do not show the results obtained by the other competing methods due to the space limit.}
\label{fig:singlestructure}
\end{figure}
$\displaystyle\textbf{Datasets.}$ The test datasets consist of $20$ image pairs: The first $10$ image pairs are single-structure datasets and they are tested in Sec.~\ref{sec:singlestructure}. The other $10$ image pairs are multiple-structure datasets, which are tested in Sec.~\ref{sec:multiplestructures}. Images, keypoint correspondences and matching scores are acquired from the BLOGS datasets\footnote{\tiny\url{http://www.cse.usf.edu/~sarkar/BLOGS/}}, the OXford VGG datasets\footnote{\tiny\url{http://www.robots.ox.ac.uk/~vgg/data/}} and the AdelaideRMF datasets\footnote{\tiny\url{http://cs.adelaide.edu.au/}}. For the image pairs in the BLOGS datasets and the OXford VGG datasets, keypoint correspondences and matching scores are not provided. We detect and match the SIFT keypoints and compute the matching scores using the VLFeat toolbox\footnote{\tiny{\url{http://www.vlfeat.org/}}}. The ground truth parameters of structures are provided by the BLOGS datasets and the OXford VGG datasets, based on which, we can manually label the ground truth correspondences. The ground truth keypoint correspondences are provided by the AdelaideRMF datasets.

\begin{table}[t]
\scriptsize
\centering
\caption{The Sampson error (and the CPU time in seconds) obtained by the competing methods for homography estimation and fundamental matrix estimation on single-structure datasets. The best results are boldfaced.}
\vspace{-0.2cm}
\label{table:singlestructure}
\begin{tabular}{|c|c|c|c|c|c|c|c|c|c|c|}
\hline
\multirow{2}{*}{} & \multicolumn{5}{c|}{Homography estimation} & \multicolumn{5}{c|}{Fundamental matrix estimation}\\
\cline{2-11}
  & Keble &  Ubc& Graffiti& Bonython & Physics & Twocars & Library & Cube& Book & Biscuit  \\
  \hline
  \hline
   Inliers (\%)& 93.85 &93.57&69.33& 26.26 & 54.71 &61.71 &70.37 & 32.11& 56.14 & 44.24 \\
  \hline
  No. of matches & 293 &140& 212& 198 & 106 & 128 & 27 & 302& 187 & 330 \\
  \hline
\hline
\multirow{2}{*}{RANSAC}& 1.71   & 0.47   & 1.32   & 10.76  & 2.44     & 0.04   & 0.01   & 0.06   & 0.03  & 0.14\\
                       & (0.29) & (0.31) & (0.30) & (6.45) & (0.43)    & (0.10) & (0.07) & (7.38) & (0.19) & (1.30)\\
\hline
\multirow{2}{*}{PROSAC}& 1.73   & 0.46   & 1.34   & 9.76   & 1.46      & 0.04   & 0.01   & 0.06   & 0.05  &0.08\\
                       & (0.33) & (0.42) & (0.35) & (6.58) & (0.66)    & (0.11) & (0.04) & (8.06) & (0.24) & (1.27)\\
\hline
\multirow{2}{*}{Astar}&{\bf1.69}& 0.47 & 1.31   &$\times$&$\times$    &{\bf0.03}&0.01  &$\times$ &$\times$  & $\times$ \\
                      & (15.25)& (1.82) & (7.23)&($>3600$)&($>3600$)  &(19.76)&(22.43)&($>3600$)&($>3600$) & ($>3600$)\\
\hline
\multirow{2}{*}{SDF}  & 1.70   &{\bf0.45}&{\bf1.28}&{\bf0.32}&{\bf1.45} &{\bf0.03}&{\bf0.00}&{\bf0.05}& {\bf0.01}&{\bf0.04}\\
                      & (0.42)  & (0.44) & (0.45) & (0.36) & (0.32)    & (2.58) & (2.70) & (0.30) & (0.29) & (0.37)\\
\hline
\end{tabular}
\end{table}
\vspace{-0.5cm}
\subsection{Single-Structure Data}
\label{sec:singlestructure}
\vspace{-0.2cm}
First, we evaluate the performance of the four fitting methods (i.e., RANSAC, PROSAC, Astar and the proposed SDF) on the 10 image pairs with single-structure data for homography estimation and fundamental matrix estimation. We report the Sampson error as \cite{litman2015inverting} (we only show the results obtained within $1$ hour as \cite{chin2015efficient}), the computational speed (i.e., the CPU time), the percentage of inliers and the number of keypoint correspondences (matches) on each image pair in Table~\ref{table:singlestructure}. For RANSAC and PROSAC, we show the average results of $50$ repeating experiments due to their randomized nature. For Astar and SDF, we do not repeat experiments due to their deterministic nature. The fitting results obtained by SDF are also shown in Fig.~\ref{fig:singlestructure}.

$\displaystyle\textbf{Homography estimation.}$ From Fig.~\ref{fig:singlestructure}(a) $\sim$ \ref{fig:singlestructure}(e) and Table~\ref{table:singlestructure}, we can see that all the four methods achieve similar Sampson errors on the image pairs with a high percentage of inliers (i.e., ``Keble", ``Ubc" and ``Graffiti"). However, for the other two image pairs with a low percentage of inliers (i.e., ``Bonython" and ``Physics"), SDF achieves the lowest Sampson errors. For the computational speed, the methods with the randomized nature (i.e., RANSAC and PROSAC) are faster than SDF on the image pairs with a high percentage of inliers. However, on the image pairs with a low percentage of inliers, SDF is significantly faster. This is because RANSAC and PROSAC cannot generate high-quality hypotheses when data contain a large number of outliers and it is difficult to determine the number of iterations to achieve the desired confidence. Furthermore, SDF shows significant superiority over the other deterministic method (i.e., Astar): Astar takes one order of magnitude more time than SDF on the image pairs with high inlier ratios (i.e., ``Keble", ``Ubc" and ``Graffiti"), and it cannot also yield results within $1$ hour on the image pairs with low inlier ratios (i.e., ``Bonython" and ``Physics"). In contrast, SDF obtains results on all image pairs within about $0.32$$\sim$$0.45$ second.

$\displaystyle\textbf{Fundamental matrix estimation.}$ From Table~\ref{table:singlestructure}, we can see that the computational speed of RANSAC and PROSAC increases as the inlier ratio decreases on all the five image pairs. As shown in Fig.~\ref{fig:singlestructure}(f) $\sim$ ~\ref{fig:singlestructure}(j) and, both methods achieve good results on the image pairs with high inlier ratios (i.e., ``Twocars", ``Library" and ``Book"). However, they are slower than SDF on ``Cube" and ``Physics" (which have low inlier ratios). Astar achieves low Sampson errors on the image pairs with high inlier ratios, but it cannot yield results within $1$ hour on the image pairs with high outlier ratios as well. SDF achieves the lowest Sampson errors on all the five image pairs, and it is much faster than the competing deterministic method (i.e., Astar).

\begin{figure}[t]
\centering
\begin{minipage}{.36\textwidth}
\centering
\centerline{\includegraphics[width=1.05\textwidth]{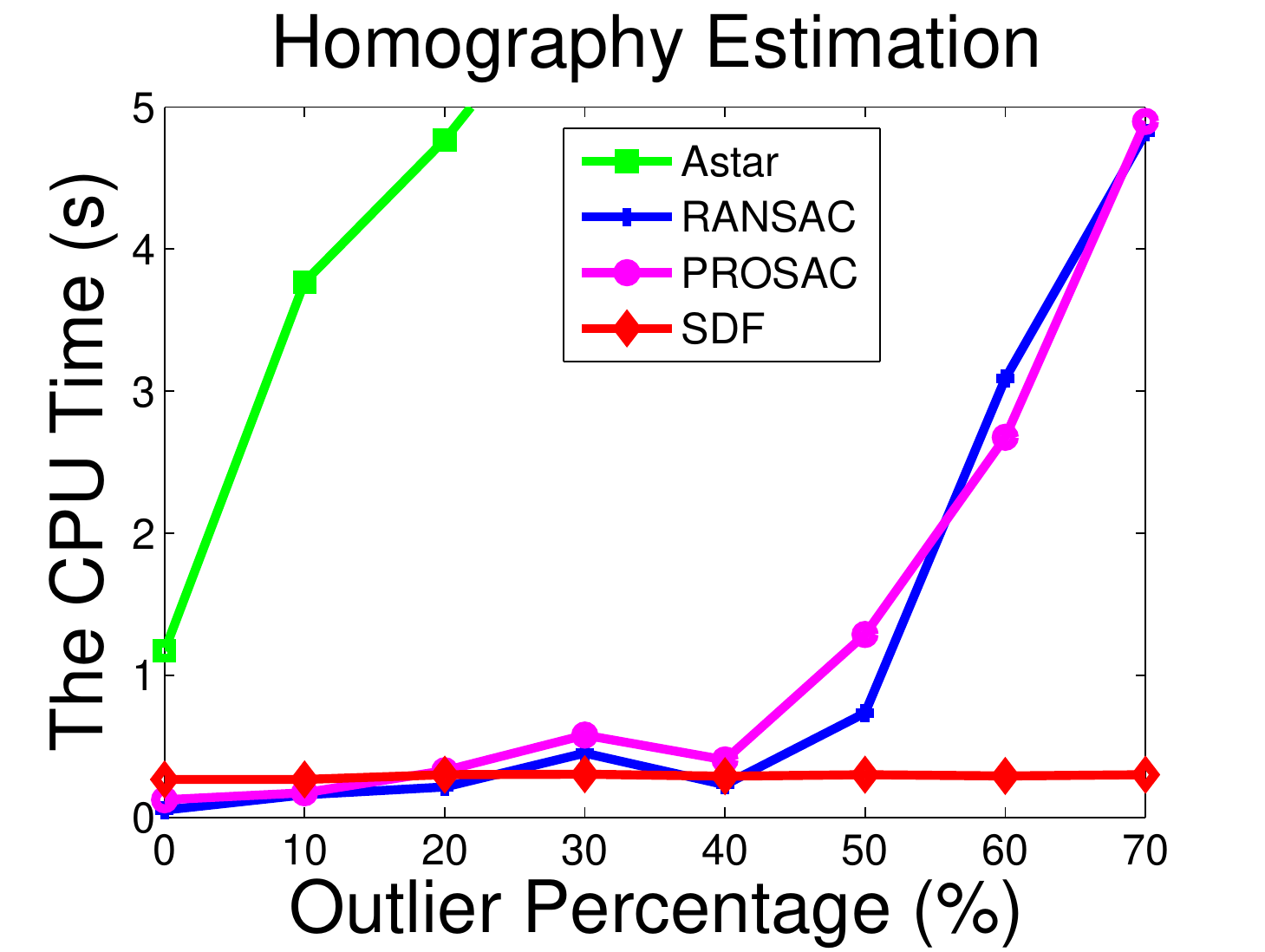}}
  \begin{center} (a) \end{center}
\end{minipage}
\begin{minipage}{.1\textwidth}
$\qquad$
\end{minipage}
\begin{minipage}{.36\textwidth}
\centerline{\includegraphics[width=1.05\textwidth]{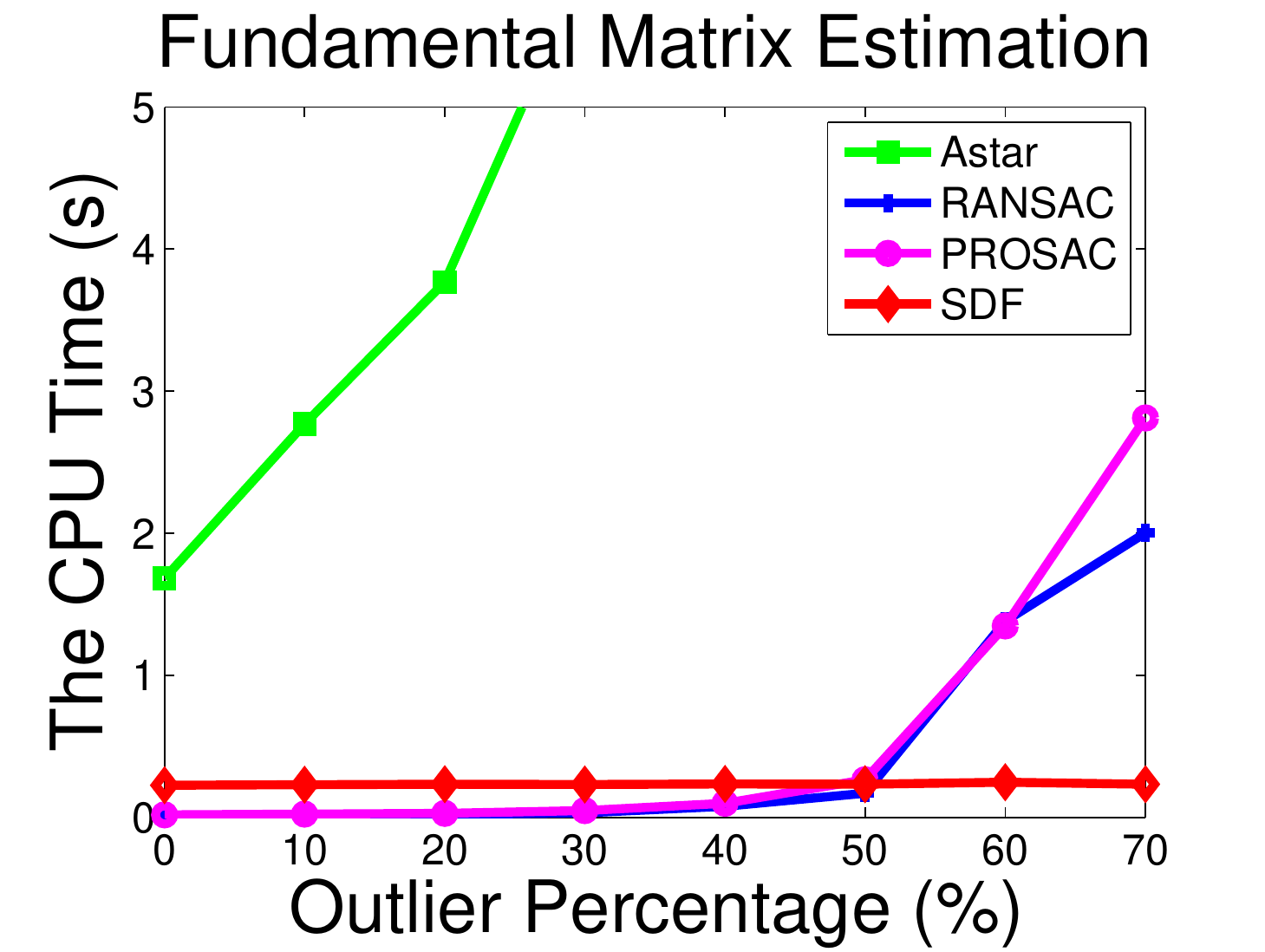}}
  \begin{center} (b) \end{center}
\end{minipage}
  \vspace{-0.4cm}
\caption{The computational speed of the four methods on two image pairs with different outlier percentages: (a) and (b) show the performance comparison on homography estimation (``Physics'') and fundamental matrix estimation (``Book"), respectively.}
\label{fig:differentoutliers}
\end{figure}
$\displaystyle\textbf{Influence of outlier percentages.}$ We also evaluate the performance of all four methods with different outlier percentages. As shown in Fig.~\ref{fig:differentoutliers}, we report the computational speed of the four competing methods on two image pairs with different outlier percentages (the Sampson errors obtained by the four methods are not reported because they are similar). We can see that SDF significantly outperforms the other three methods when the outlier percentage is larger than $50\%$. The CPU time of Astar is much higher than that of RANSAC, PROSAC and SDF. The CPU time of RANSAC and PROSAC increases substantially when the outlier percentage is larger than $40\%$ (for homography estimation)  and $50\%$ (for fundamental matrix estimation). This is because they suffer from the influence of outliers during the process of generating all-inlier subsets. In contrast, the CPU time of SDF has no significant change on both image pairs when the outlier percentage increases, which shows the robustness of SDF to outliers.

\begin{figure}[t]
\centering
\begin{minipage}{.36\textwidth}
\centering
\centerline{\includegraphics[width=1.05\textwidth]{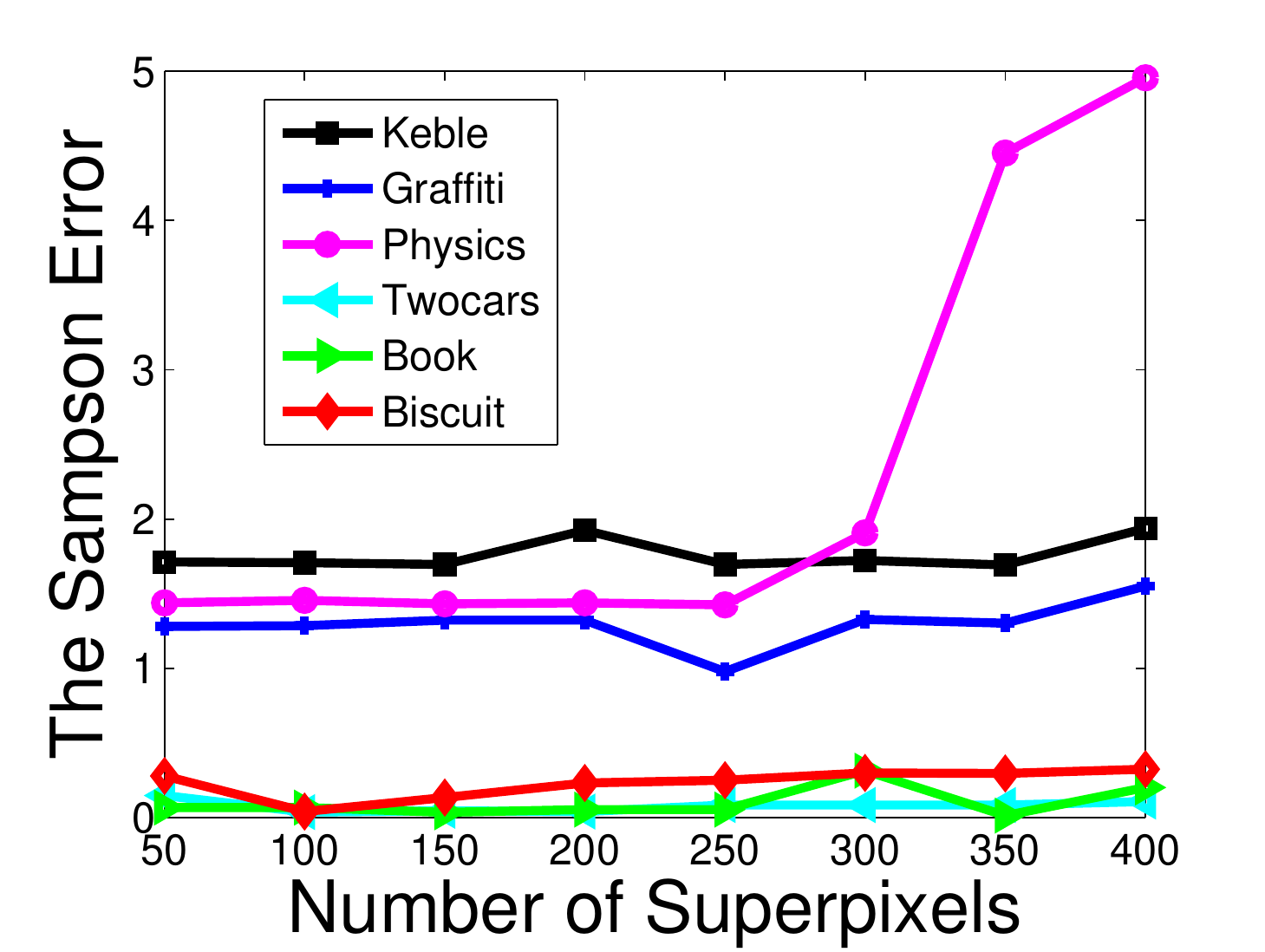}}
  \begin{center} (a) \end{center}
\end{minipage}
\begin{minipage}{.1\textwidth}
$\qquad$
\end{minipage}
\begin{minipage}{.36\textwidth}
\centerline{\includegraphics[width=1.05\textwidth]{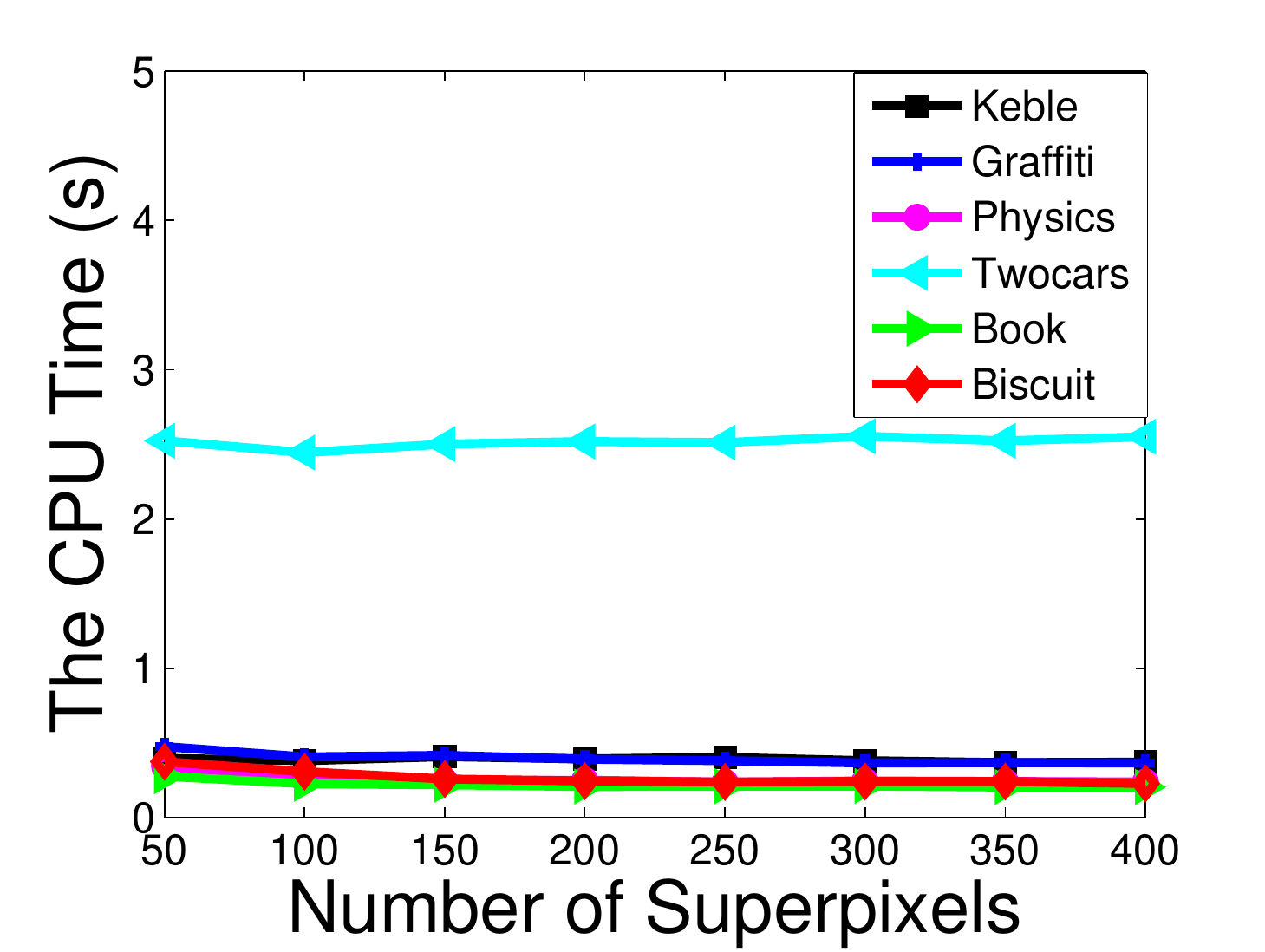}}
  \begin{center} (b) \end{center}
\end{minipage}
  \vspace{-0.4cm}
\caption{The results of the proposed SDF with different numbers of superpixels on six image pairs: (a) and (b) show the performance comparison on the Sampson error and the computational speed, respectively.}
\label{fig:differentsuperpixels}
\end{figure}

$\displaystyle\textbf{Influence of the number of superpixels.}$ Note that, compared with RANSAC and PROSAC, SDF uses an extra parameter for superpixel segmentation, i.e., the number of superpixels ($M$). Thus, we test the influence of the number of superpixels on the performance of SDF. As shown in Fig.~\ref{fig:differentsuperpixels}, we show the Sampson error and the computational speed of SDF with different numbers of superpixels on six image pairs (three image pairs, i.e., ``Keble", ``Graffiti" and ``Physics", for homography estimation, and the other three image pairs, i.e., ``Twocars", ``Book" and ``Biscuit", for fundamental matrix estimation). We can see that SDF consistently achieves low Sampson errors on five out of the six image pairs with different numbers of superpixels. However, the Sampson error obtained by SDF dramatically increases on ``Physics" when $M$ is larger than $300$. The reason behind this is that ``Physics" includes few inliers and each group partitioned by SDF includes much less inliers as $M$ becomes larger than $300$. In such a case, sampling an all-inlier subset from each group is difficult, and the quality of hypotheses generated by the sampled subsets will affect the fitting results of SDF.

For the computational speed, the result of SDF does not change a lot on all image pairs as $M$ increases. SDF takes more CPU time on ``Twocars" than the other five image pairs due to the complex scenario of this image pair (which affects the computational speed of SLIC for superpixel generation). Therefore, we experimentally set the number of superpixels within the range of $[50, 300]$.

\begin{figure}
\centering
\begin{minipage}[t]{.421\textwidth}
  \centering
  \centerline{\includegraphics[width=1.18\textwidth]{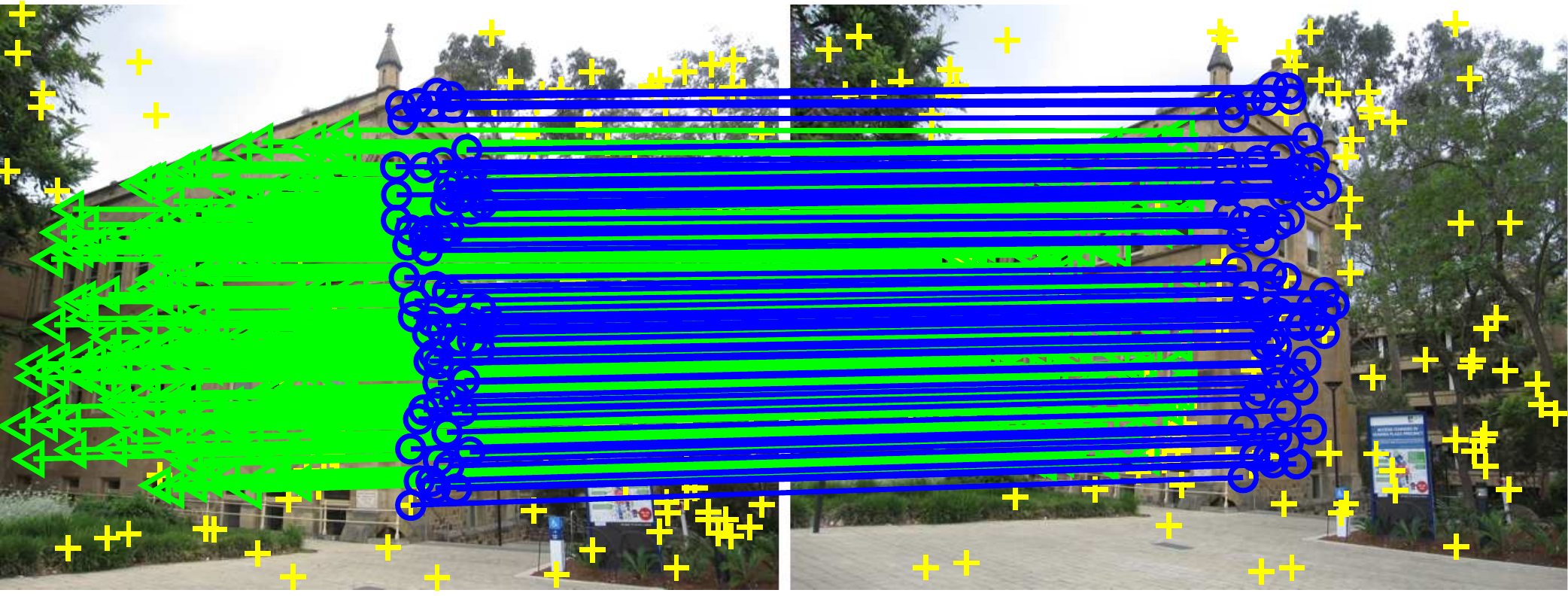}}
  \centerline{Inliers (\%): 18.73 and 48.81}
  \centerline{(a) Oldclassicswing (379 matches: 2 models)}
  \centerline{\includegraphics[width=1.18\textwidth]{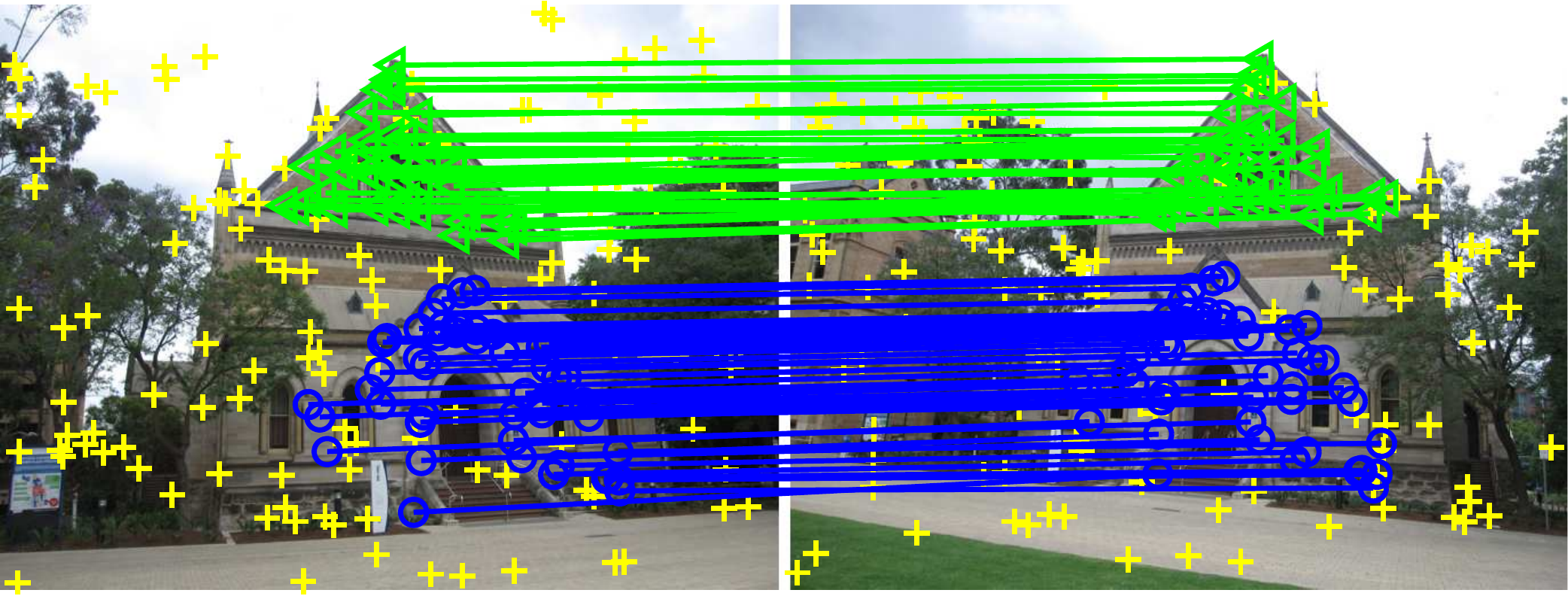}}
  \centerline{Inliers: 17.76 and 21.50}
    \centerline{(b) Elderhalla (214 matches: 2 models)}
    \centerline{\includegraphics[width=1.18\textwidth]{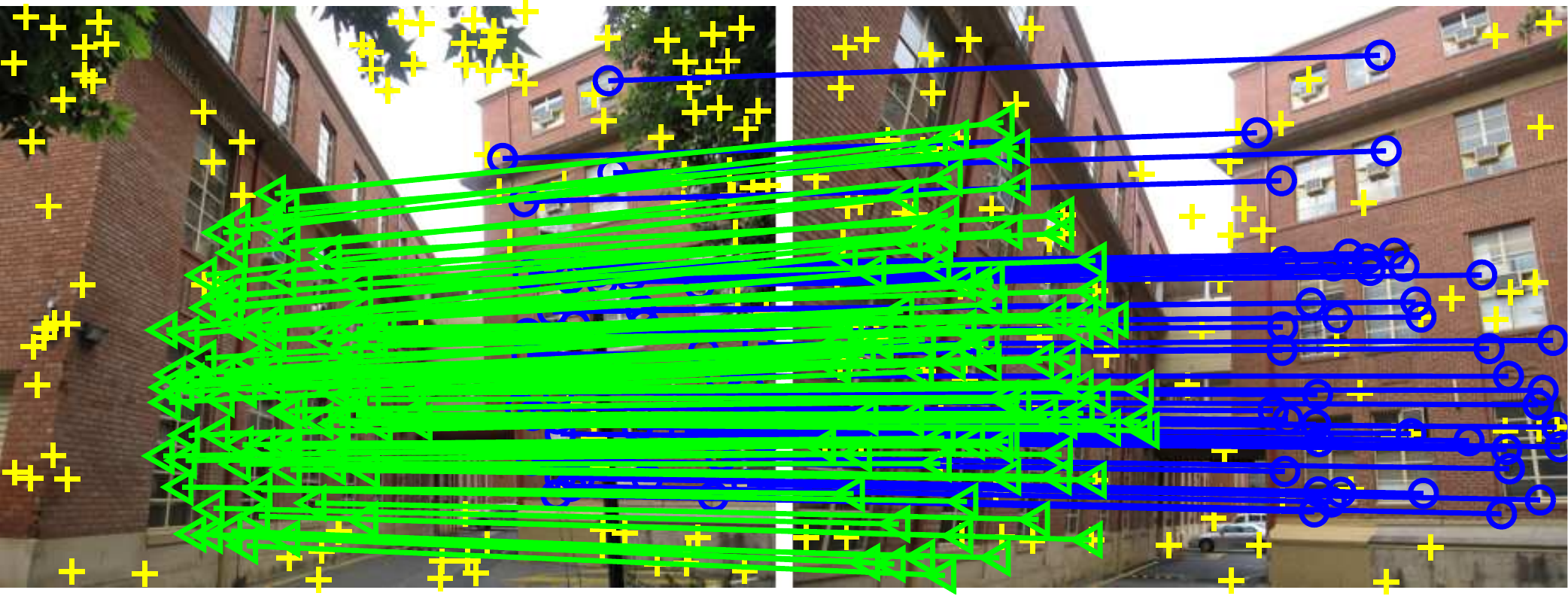}}
   \centerline{Inliers (\%): 18.40 and 34.40}
  \centerline{(c) Sene (250 matches: 2 models)}
  \centerline{\includegraphics[width=1.18\textwidth]{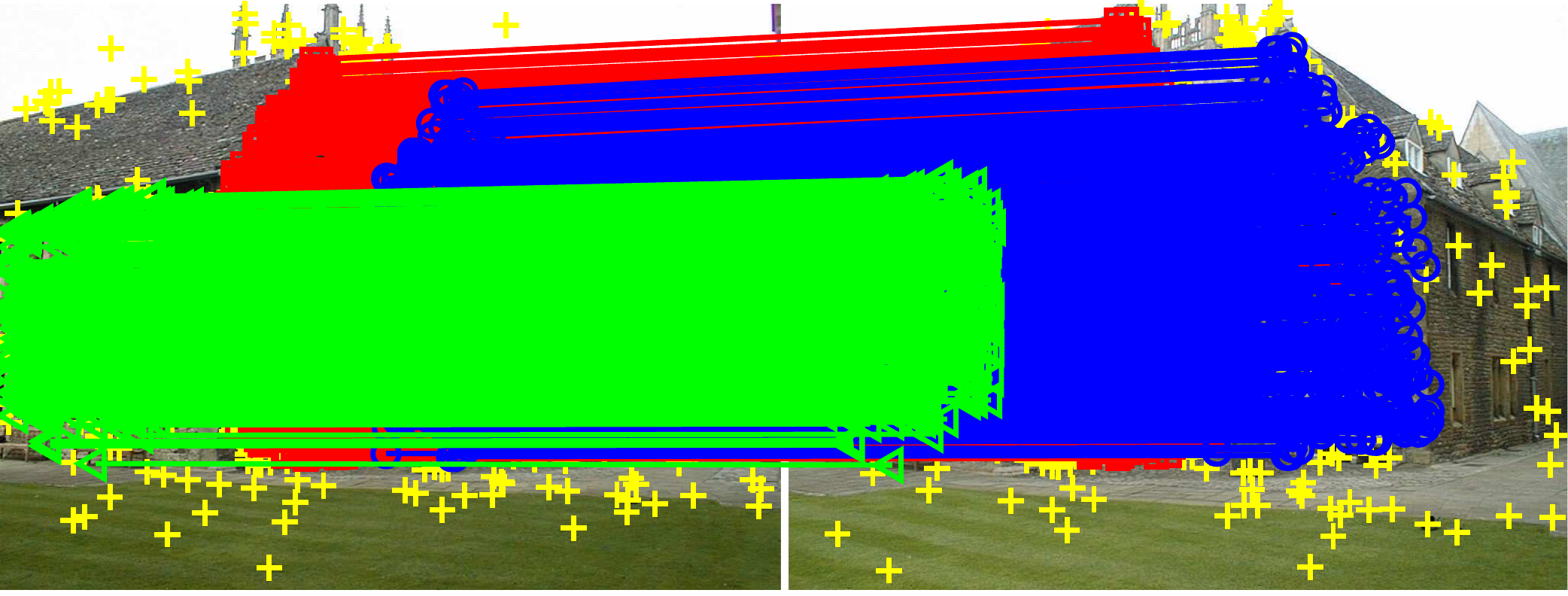}}
  \centerline{Inliers (\%): 28.17, 28.17 and 28.17}
    \centerline{(d) MC3 (1775 matches: 3 models)}
      \centerline{\includegraphics[width=1.18\textwidth]{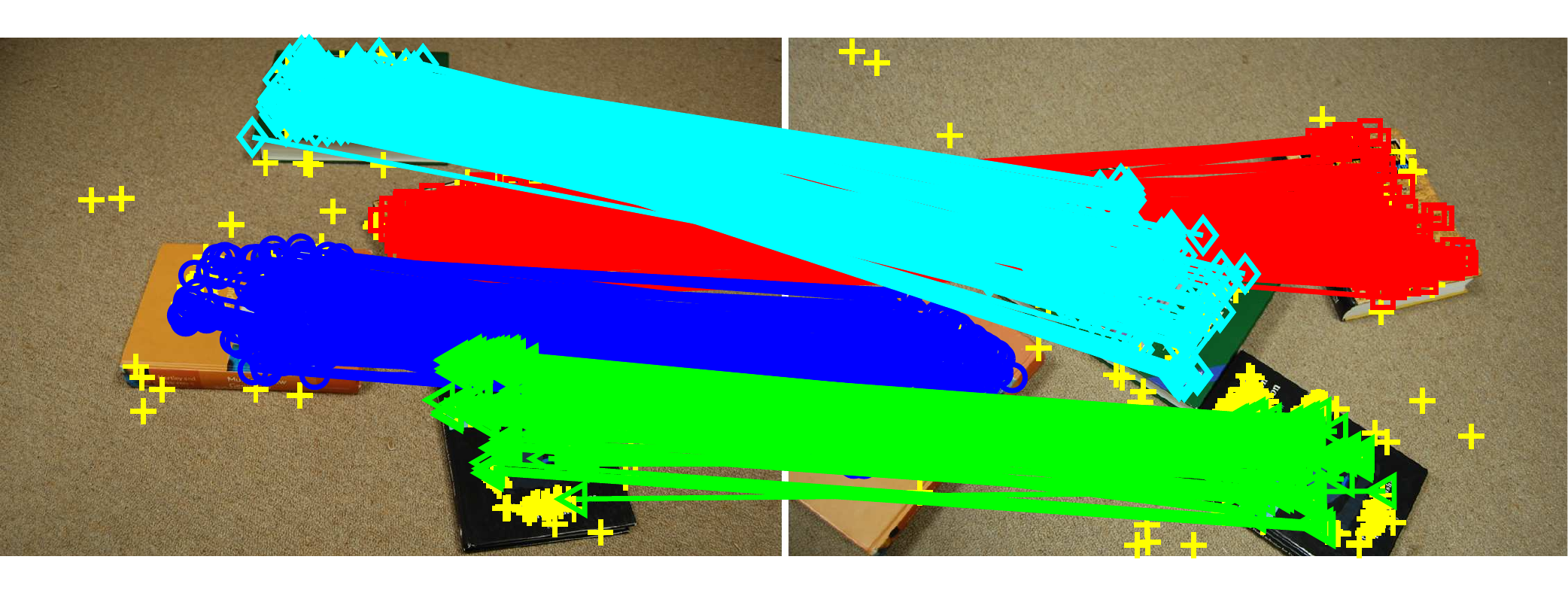}}
   \centerline{Inliers (\%): 15.70, 16.09, 17.50 and 29.73}
    \centerline{(e) 4B (777 matches: 4 models)}
\end{minipage}
\begin{minipage}{.1\textwidth}
$\qquad$
\end{minipage}
\begin{minipage}[t]{.421\textwidth}
  \centering
  \centerline{\includegraphics[width=1.18\textwidth]{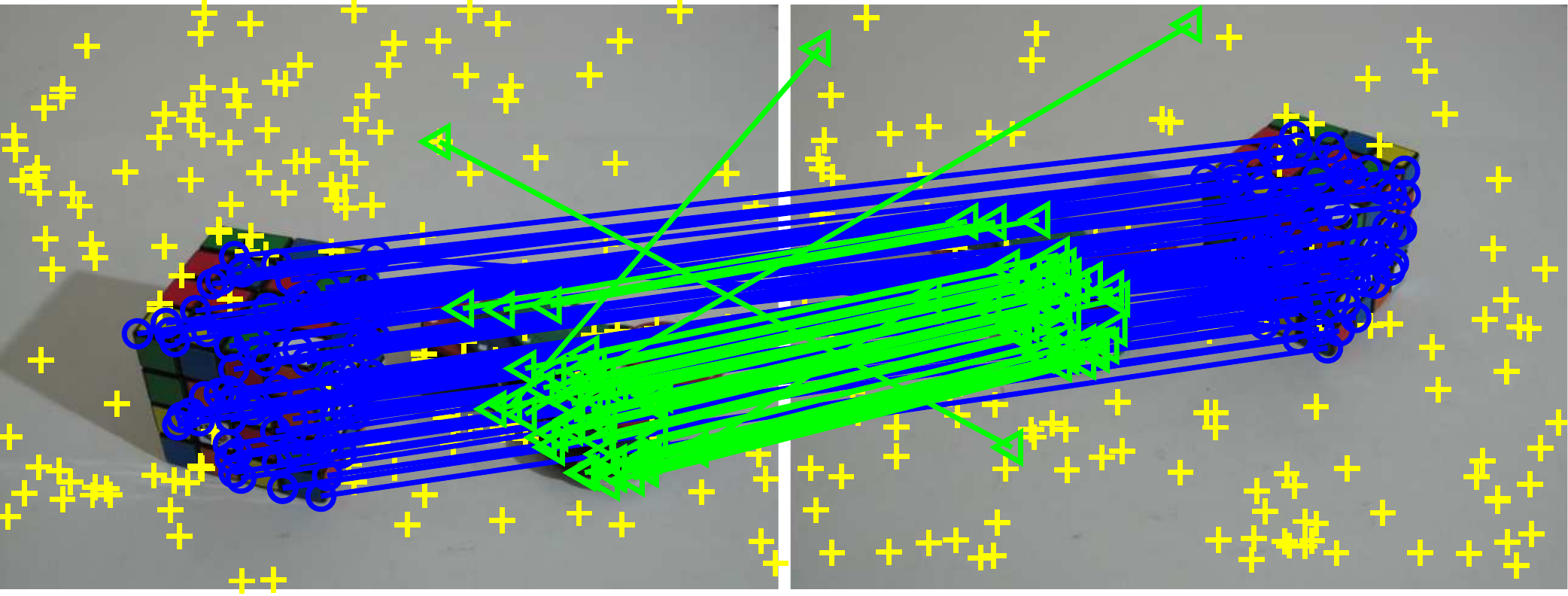}}
  \centerline{Inliers (\%):20.07 and 29.58}
  \centerline{(f) Cubechips (284 matches: 2 models)}
  \centerline{\includegraphics[width=1.18\textwidth]{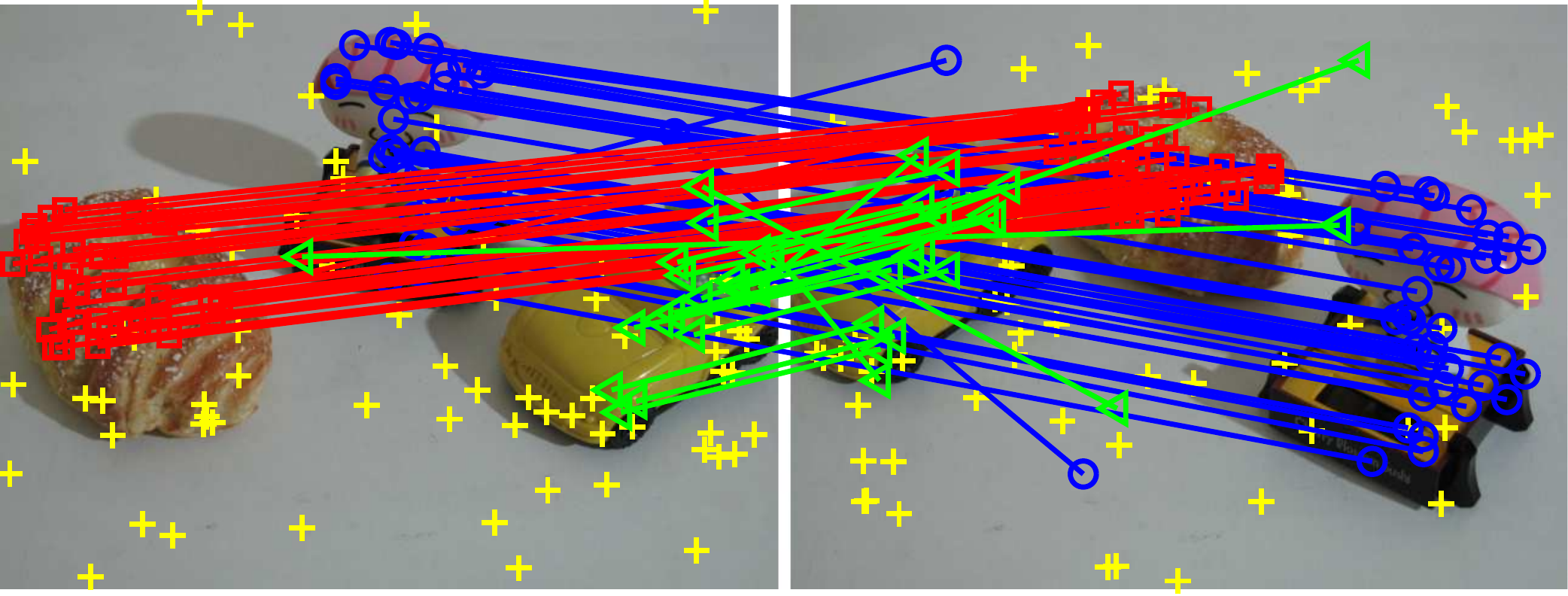}}
  \centerline{Inliers (\%): 20.48, 23.49 and 22.29}
  \centerline{(g) Breadtoycar (166 matches: 3 models)}
  \centerline{\includegraphics[width=1.18\textwidth]{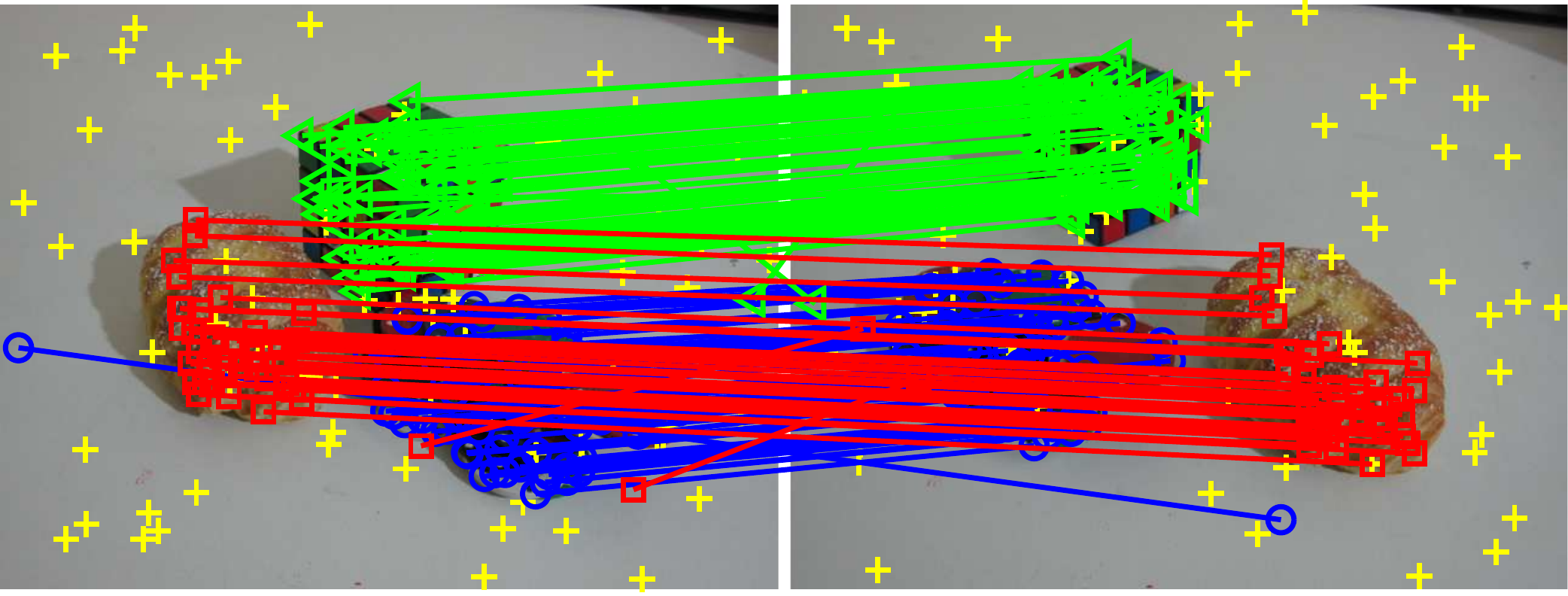}}
  \centerline{Inliers (\%):14.78, 24.78 and 25.22}
  \centerline{(h) Breadcubechips (230 matches: 3 models)}
  \centerline{\includegraphics[width=1.18\textwidth]{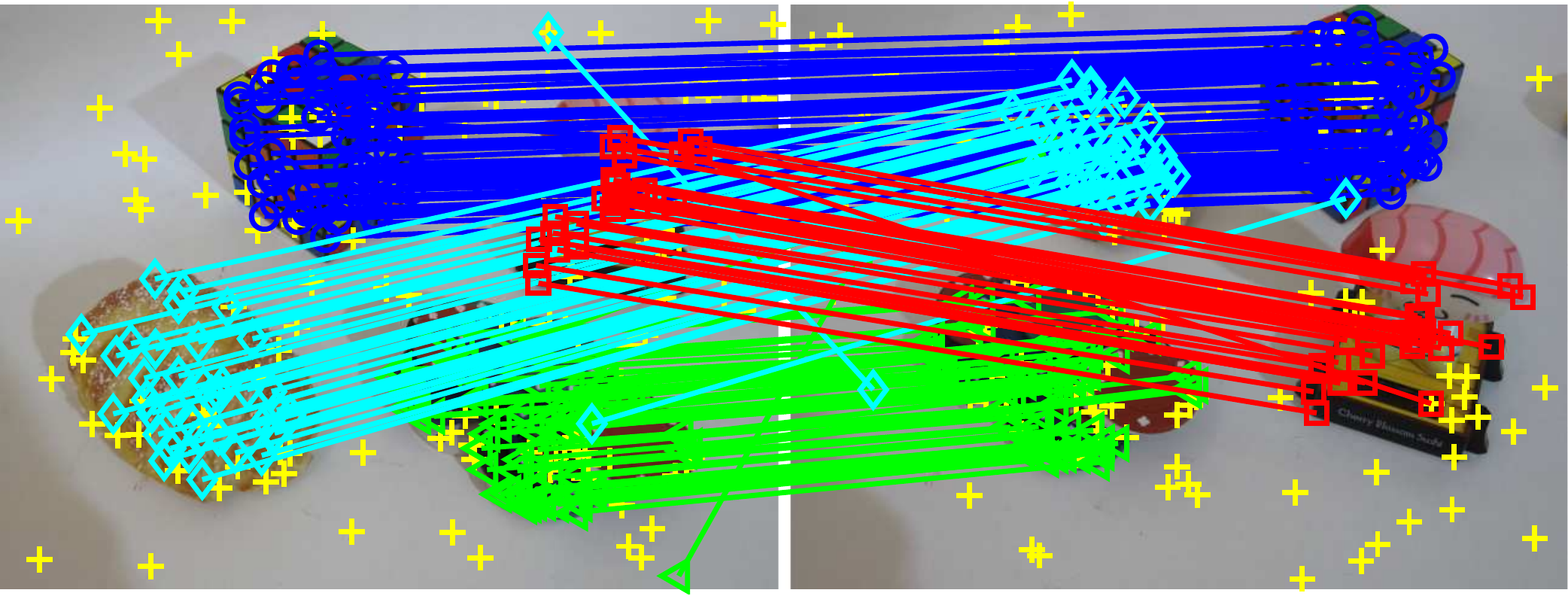}}
  \centerline{Inliers (\%): 11.62, 14.98, 21.71 and 24.77}
    \centerline{(i) Cubebreadtoychips (327 matches: 4 models)}
      \centerline{\includegraphics[width=1.18\textwidth]{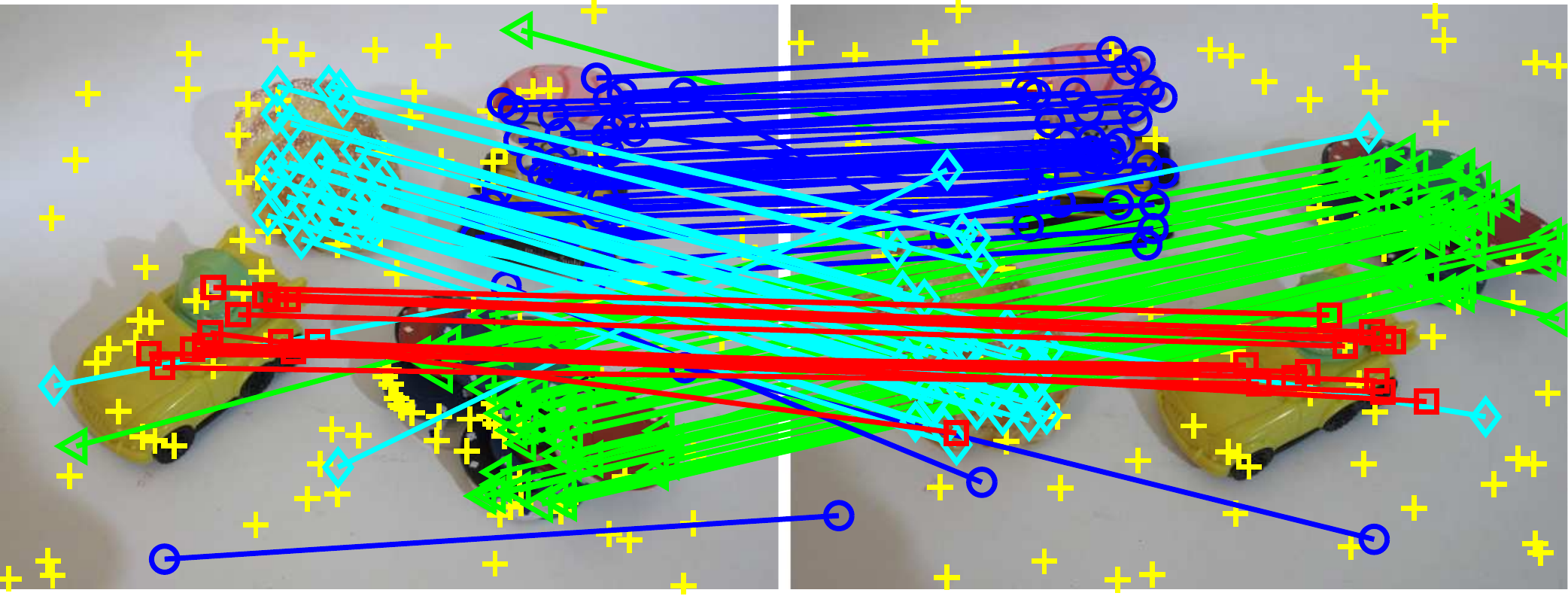}}
      \centerline{Inliers (\%): 9.70, 13.92, 17.30 and 24.47}
    \centerline{(j) Breadcartoychips (237 matches: 4 models)}
\end{minipage}
\caption{Fitting results obtained by SDF on 10 image pairs with multiple-structure data. (a)$\sim$(e) show the results obtained by SDF on homography estimation, and (f)$\sim$(j) show the results obtained by SDF on fundamental matrix estimation. We also report the percentage of inliers for each model instance, the number of model instances and the number of keypoint correspondences (matches) on each image pair. We do not show the results obtained by the other competing methods due to the space limit.}
\label{fig:multiplestructure}
\end{figure}
\vspace{-0.5cm}
\subsection{Multiple-Structure Data}
\label{sec:multiplestructures}
   \vspace{-0.2cm}
\begin{table}[b]
\centering
\scriptsize
\caption{The Sampson error (and the CPU time in seconds) obtained by the competing methods for homography estimation and fundamental matrix estimation on multiple-structure datasets. The best results are boldfaced.}
\vspace{-0.2cm}
\begin{tabular}{|c|c|c|c|c|c|c|c|c|c|c|}
\hline
\multirow{2}{*}{} & \multicolumn{5}{c|}{Homography estimation} & \multicolumn{5}{c|}{Fundamental matrix estimation}\\
\cline{2-11}
  &D1& D2 &D3&D4&D5&D6&D7&D8&D9&D10\\
  \hline
\multirow{2}{*}{RANSAC}&1.26& 1.12  & 0.78   & 3.76   & 1.55 & 0.04   & 0.80   & 0.41   & 0.09  & 4.13\\
                       &(1.33)& (57.70) & (11.15) & (3.83) &(3.22) & (20.53) & (21.08)&(23.96) & (37.18) & (27.71)\\
\hline
\multirow{2}{*}{PROSAC}&2.73& 1.11 & 0.87  & $\times$   & 1.48  & 0.09   & 0.22   & 0.29   & 0.47  &6.78\\
                       &(2.76)&(33.25)& (19.37) & ($>3600$) & (7.84)  & (20.92) & (21.39) &(23.92) & (37.81) & (27.97)\\
\hline
\multirow{2}{*}{T-linkage}&{\bf1.02}&{\bf1.09}&0.76& 3.75 &{\bf1.36}&{\bf0.03}&0.13 & {0.12} &0.11 & 0.13 \\
                      &(62.21)& (23.96)&(30.91)&(1508.62)&(233.02)&(45.05)&(22.94)&(31.84)&(53.38) & (33.88)\\
\hline
\multirow{2}{*}{SDF}  &1.03& 1.11&{\bf0.75}&{\bf3.73}&{1.47} &{\bf0.03}&{\bf0.11}&{\bf0.10}& {\bf0.07}&{\bf0.04}\\
                      &(0.84)& (0.72) & (0.46) &(1.96) & (1.51)  & (0.64) & (0.54) & (0.61) & (0.71) & (0.63)\\

\hline
\end{tabular}
\\
\medskip
(D1-Oldclassicswing; D2-Elderhalla; D3-Sene; D4-MC3; D5-4B; D6-Cubechips; D7-Breadtoycar; D8-Breadcubechips; D9-Cubebreadtoychips; D10-Breadcartoychips.)
\label{table:multiplestructure}
\end{table}
In this subsection, we evaluate the performance of the four fitting methods (i.e., RANSAC, PROSAC, T-linkage and SDF) on 10 image pairs with multiple-structure data for homography estimation and fundamental matrix estimation. We report the Sampson errors obtained by the competing methods (we only show the results obtained within $1$ hour) and their computational speed in Table~\ref{table:multiplestructure}. For RANSAC, PROSAC and T-linkage, we show the average results of $50$ repeating experiments due to their randomized nature. The fitting results obtained by SDF are also shown in Fig.~\ref{fig:multiplestructure}.

$\displaystyle\textbf{Homography estimation.}$ From Fig.~\ref{fig:multiplestructure}(a) $\sim$ \ref{fig:multiplestructure}(e) and Table~\ref{table:multiplestructure}, we can see that RANSAC achieves low Sampson errors on all five image pairs, but it is much slower than SDF. This is more obvious on the image pairs with low inlier ratios (e.g., SDF is about $80.13$ times faster than RANSAC on ``Elderhalla''). PROSAC succeeds in fitting four image pairs, but it fails in fitting ``MC3'' since it cannot sample non-degenerate subsets within $1$ hour. T-linkage achieves the lowest Sampson errors on $3$ out of $5$ image pairs, but it suffers from high computational complexity due to a large number of keypoint correspondences used during the agglomerative clustering procedure (e.g., SDF is $769.70$ times faster than T-linkage on ``MC3''). In contrast, SDF achieves the fastest computational speed among the four fitting methods for all five image pairs: SDF is about $1.58$$\sim$$80.14$ times faster than RANSAC, and it is about $3.2$ times $\sim$ three order of magnitude faster than PROSAC, and about one to two order of magnitude faster than T-linkage. 

$\displaystyle\textbf{Fundamental matrix estimation.}$ From Fig.~\ref{fig:multiplestructure}(f) $\sim$ \ref{fig:multiplestructure}(j) and Table~\ref{table:multiplestructure}, we can see that both RANSAC and PROSAC need much time to determine the number of iterations to achieve high confidence on all five image pairs. This is because that it is difficult for them to sample all-inlier subsets when a dataset has a high outlier percentage and the estimation needs to sample a large number of subsets. T-linkage achieves lower Sampson errors than RANSAC and PROSAC due to its robustness to outliers. However, SDF achieves the lowest Sampson error for all the five image pairs, and it is much faster than the other three fitting methods (about $32.08$$\sim$$52.37$ times faster than RANSAC, and about $32.69$$\sim$$53.25$ times faster than PROSAC, and $42.48$$\sim$$75.18$ times faster than T-linkage). The results show the effectiveness of SDF for fitting multiple-structure data

\section{Conclusions}
\label{sec:conclusion}
In this paper, we propose a simple but effective determinstic fitting method (SDF) that introduces prior information of feature appearance to geometric model fitting. We show that prior information of feature appearance (derived from superpixels) can provide powerful grouping cues for the proposed deterministic sampling algorithm to generate consistent hypotheses, where keypoint correspondences with high matching scores are selected as sampled subsets. The generated hypotheses contain a high percentage of good hypotheses with a small percentage of bad hypotheses. Based on the advantage of the generated hypotheses, a novel fit-and-remove framework is proposed for model selection. SDF effectively combines hypothesis generation and model selection to deterministically deal with the two-view model fitting problems.

Compared with the fitting methods with randomized nature (e.g., RANSAC and T-linkage), SDF is tractable and can deterministically provide consistent solutions for model fitting. Compared with the feature appearance based fitting methods (e.g., PROSAC), SDF can obtain better performance in both speed and accuracy. SDF has also significant superiority over several other deterministic methods (e.g., BnB~\cite{li2009consensus} and Astar~\cite{chin2015efficient}): SDF is much faster, and it can achieve promising performance on image pairs with both single-structure and multiple-structure data.
\section*{Acknowledgment}
\small{This work was supported by the National Natural Science Foundation of China under Grants 61472334 and 61571379. David Suter acknowledged funding under ARC DPDP130102524.}
\clearpage

\end{document}